\pdfoutput=1

\documentclass[runningheads]{llncs}

\usepackage{eccv}

\usepackage{eccvabbrv}

\usepackage{graphicx}
\usepackage{booktabs}
\usepackage{tikz}
\usetikzlibrary{spy}
\usepackage{multirow}
\usepackage{subcaption}
\usepackage{overpic}
\usepackage{marvosym}

\usepackage{booktabs}
\usepackage{adjustbox}
\usepackage{xcolor}    
\usepackage{tabulary}
\usepackage{colortbl}
\usepackage{makecell}
\usepackage{float}
\usepackage{url}
\usepackage{balance}
\usepackage[utf8]{inputenc} % allow utf-8 input
\usepackage[T1]{fontenc}    % use 8-bit T1 fonts
\usepackage{wrapfig}
\definecolor{color3}{rgb}{0.95,0.95,0.95}

\usepackage{graphicx}    
\usepackage{pgfplots}
\usepackage{bchart}
\usetikzlibrary{decorations.pathreplacing}
\usetikzlibrary{patterns}
\pgfplotsset{compat=1.16} 

\usepackage[accsupp]{axessibility}  % Improves PDF readability for those with disabilities.

\usepackage{hyperref}

\usepackage{orcidlink}

\begin{document}

\title{HiT-SR: Hierarchical Transformer for Efficient Image Super-Resolution} 


% Second Author\inst{2,3}\orcidlink{1111-2222-3333-4444} \and
% Third Author\inst{3}\orcidlink{2222--3333-4444-5555}}
\author{Xiang Zhang\inst{1}\orcidlink{0000-0003-2004-5794} \and
Yulun Zhang\inst{2}\thanks{Corresponding author: Yulun Zhang.}\orcidlink{0000-0002-2288-5079} \and
Fisher Yu\inst{1}\orcidlink{0000-0001-8829-7344}}


% Springer Heidelberg, Tiergartenstr.~17, 69121 Heidelberg, Germany
\institute{ETH Zürich, Switzerland \and
MoE Key Lab of Artificial Intelligence, Shanghai Jiao Tong University, China\\
\email{\{xiangz.ethz,yulun100\}@gmail.com, i@yf.io}\\
\url{https://github.com/XiangZ-0/HiT-SR}}
\maketitle

% \vspace{-3mm}
\begin{abstract}
    Transformers have exhibited promising performance in computer vision tasks including image super-resolution (SR). However, popular transformer-based SR methods often employ window self-attention with quadratic computational complexity to window sizes, resulting in fixed small windows with limited receptive fields. In this paper, we present a general strategy to convert transformer-based SR networks to hierarchical transformers (HiT-SR), boosting SR performance with multi-scale features while maintaining an efficient design. Specifically, we first replace the commonly used fixed small windows with expanding hierarchical windows to aggregate features at different scales and establish long-range dependencies. Considering the intensive computation required for large windows, we further design a spatial-channel correlation method with linear complexity to window sizes, efficiently gathering spatial and channel information from hierarchical windows. Extensive experiments verify the effectiveness and efficiency of our HiT-SR, and our improved versions of SwinIR-Light, SwinIR-NG, and SRFormer-Light yield state-of-the-art SR results with fewer parameters, FLOPs, and faster speeds ($\sim7\times$). 
    % Code, models, and results are available at \url{https://github.com/XiangZ-0/HiT-SR}.
\end{abstract}    
% \vspace{-3mm}
\section{Introduction}
% \vspace{-2mm}
Image super-resolution (SR) is a classical low-level vision task that aims to convert a low-resolution (LR) image to its high-resolution (HR) counterpart with better visual details. How to tackle the ill-posed SR problem has attracted considerable interest for decades \cite{freeman2000learning,yang2008image,glasner2009super,yang2010image,srcnn_dong2014learning,hu2019channel}. Many popular approaches employ convolutional neural networks (CNNs) to learn the projection between LR inputs and HR images \cite{srcnn_dong2014learning,rcan_zhang2018image,dai2019second,zhang2019residual,mei2020image,niu2020single}. Despite significant progress being achieved, CNN-based methods usually focus on utilizing local features via convolution and often fall short in aggregating long-range information across the image, limiting the performance of CNN-based SR.

\def\scale{0.3}
\def\efficiencyscale{0.3}

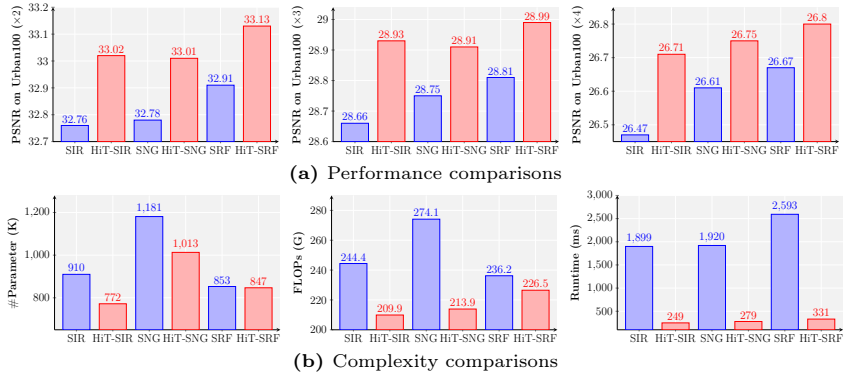
\begin{figure}
    \centering
\begin{subfigure}{\linewidth}
\centering
%% for performance ==========================
% for params
\begin{tikzpicture}[scale=\scale]
       % \draw[help lines] (0, 0)  grid (5,5); 
		\begin{axis}
			[ybar,
			 symbolic x coords={SIR, HiT-SIR, SNG, HiT-SNG, SRF, HiT-SRF},
              ymin=32.7, ymax=33.2,
              height=0.618\textwidth,
              width=0.95\textwidth,
              ylabel={\textbf{PSNR on Urban100 ($\times 2$)}},
              line width=1.2pt,
              bar width=1.2cm,
			 axis y line=left,
			 axis x line=bottom,
              nodes near coords, 
              every axis/.append style={font=\Large},
              grid,
    grid style={color=white, line width=0.5pt},
              axis background/.style={fill=gray!10}, % 设置背景颜色
              bar shift=0cm,
			 enlarge x limits=0.12, ] 
			\addplot+ coordinates {(SIR, 32.76)  (SNG, 32.78) (SRF, 32.91)}; 
			\addplot+ coordinates {(HiT-SIR, 33.02)  (HiT-SNG, 33.01) (HiT-SRF, 33.13)}; 
		\end{axis} 
	\end{tikzpicture} 
  % for FLOPs
    \begin{tikzpicture}[scale=\scale]
       % \draw[help lines] (0, 0)  grid (5,5); 
		\begin{axis}
			[ybar,
			 symbolic x coords={SIR, HiT-SIR, SNG, HiT-SNG, SRF, HiT-SRF},
              height=0.618\textwidth,
              width=0.95\textwidth,
              ymin=28.6, ymax=29.04,
              ylabel={\textbf{PSNR on Urban100 ($\times 3$)}},
              line width=1.2pt,
              bar width=1.2cm,
			 axis y line=left,
			 axis x line=bottom,
              nodes near coords, 
              every axis/.append style={font=\Large},
              grid,
    grid style={color=white, line width=0.5pt},
              axis background/.style={fill=gray!10}, % 设置背景颜色
              bar shift=0cm,
			 enlarge x limits=0.12, ] 
			\addplot+ coordinates {(SIR, 28.66)  (SNG, 28.75) (SRF, 28.81)}; 
			\addplot+ coordinates {(HiT-SIR, 28.93) (HiT-SNG, 28.91) (HiT-SRF, 28.99) }; 
		\end{axis} 
	\end{tikzpicture} 
 % for runtime
 \begin{tikzpicture}[scale=\scale]
       % \draw[help lines] (0, 0)  grid (5,5); 
		\begin{axis}
			[ybar,
			 symbolic x coords={SIR, HiT-SIR, SNG, HiT-SNG, SRF, HiT-SRF},
              height=0.618\textwidth,
              width=0.95\textwidth,
              ymin=26.45, ymax=26.85,
              ylabel={\textbf{PSNR on Urban100 ($\times 4$)}},
              line width=1.2pt,
              bar width=1.2cm,
			 axis y line=left,
			 axis x line=bottom,
              nodes near coords, 
              every axis/.append style={font=\Large},
              grid,
    grid style={color=white, line width=0.5pt},
              axis background/.style={fill=gray!10}, % 设置背景颜色
              bar shift=0cm,
			 enlarge x limits=0.12, ]  
			\addplot+ coordinates {(SIR, 26.47)  (SNG, 26.61) (SRF, 26.67)}; 
			\addplot+ coordinates {(HiT-SIR, 26.71) (HiT-SNG, 26.75) (HiT-SRF, 26.80)}; 
		\end{axis} 
	\end{tikzpicture} 
    \caption{Performance comparisons}
\end{subfigure}
\\
\begin{subfigure}{\linewidth}
\centering
%% for complexity ===========================
     % for params
       \begin{tikzpicture}[scale=\efficiencyscale]
       % \draw[help lines] (0, 0)  grid (5,5); 
		\begin{axis}
			[ybar,
			 symbolic x coords={SIR, HiT-SIR, SNG, HiT-SNG, SRF, HiT-SRF},
            height=0.618\textwidth,
              width=0.95\textwidth,
              ymin=650, ymax=1280,
              ylabel={\textbf{\#Parameter (K)}},
              line width=1.2pt,
              bar width=1.2cm,
			 axis y line=left,
			 axis x line=bottom,
              nodes near coords, 
              every axis/.append style={font=\Large},
              grid,
    grid style={color=white, line width=0.5pt},
              axis background/.style={fill=gray!10}, % 设置背景颜色
              bar shift=0cm,
			 enlarge x limits=0.12, ] 
			\addplot+ coordinates {(SIR, 910)  (SNG, 1181) (SRF, 853)}; 
			\addplot+ coordinates {(HiT-SIR, 772) (HiT-SNG, 1013) (HiT-SRF, 847)}; 
		\end{axis} 
	\end{tikzpicture} 
  % for FLOPs
    \begin{tikzpicture}[scale=\efficiencyscale]
       % \draw[help lines] (0, 0)  grid (5,5); 
		\begin{axis}
			[ybar,
			 symbolic x coords={SIR, HiT-SIR, SNG, HiT-SNG, SRF, HiT-SRF},
            height=0.618\textwidth,
              width=0.95\textwidth,
              ymin=200, ymax=290,
              ylabel={\textbf{FLOPs (G)}},
              line width=1.2pt,
              bar width=1.2cm,
			 axis y line=left,
			 axis x line=bottom,
              nodes near coords, 
              every axis/.append style={font=\Large},
              grid,
    grid style={color=white, line width=0.5pt},
              axis background/.style={fill=gray!10}, % 设置背景颜色
              bar shift=0cm,
			 enlarge x limits=0.12, ] 
			\addplot+ coordinates {(SIR, 244.4)  (SNG, 274.1) (SRF, 236.2)}; 
			\addplot+ coordinates {(HiT-SIR, 209.9) (HiT-SNG, 213.9) (HiT-SRF, 226.5)}; 
		\end{axis} 
	\end{tikzpicture} 
 % for runtime
 \begin{tikzpicture}[scale=\efficiencyscale]
       % \draw[help lines] (0, 0)  grid (5,5); 
		\begin{axis}
			[ybar,
			 symbolic x coords={SIR, HiT-SIR, SNG, HiT-SNG, SRF, HiT-SRF},
             height=0.618\textwidth,
              width=0.95\textwidth,
              ymin=100, ymax=3000,
              ylabel={\textbf{Runtime (ms)}},
              line width=1.2pt,
              bar width=1.2cm,
			 axis y line=left,
			 axis x line=bottom,
              nodes near coords, 
              every axis/.append style={font=\Large},
              grid,
    grid style={color=white, line width=0.5pt},
              axis background/.style={fill=gray!10}, % 设置背景颜色
              bar shift=0cm,
			 enlarge x limits=0.12, ] 
			\addplot+ coordinates {(SIR, 1899)  (SNG, 1920) (SRF, 2593)}; 
			\addplot+ coordinates {(HiT-SIR, 249) (HiT-SNG, 279) (HiT-SRF, 331)}; 
		\end{axis} 
	\end{tikzpicture} 
    \caption{Complexity comparisons}
    \label{fig:first-efficiency}
\end{subfigure}
    % \vspace{-2.1em}
    \caption{Comparisons of
    the popular efficient SR transformers, \ie, SwinIR-Light (SIR) \cite{liang2021swinir}, SwinIR-NG (SNG) \cite{choi2023ngram}, and SRFormer-Light (SRF) \cite{Zhou_2023srformer}, and the corresponding HiT-SR versions, \ie, HiT-SIR, HiT-SNG, and HiT-SRF. The complexity metrics are calculated under $\times 2$ upscaling on an A100 GPU, with the output size set to $720 \times 1280$. }
    \label{fig:first}
    % \vspace{-2em}
\end{figure}

\par 
The recent development of vision transformers provides a promising solution for establishing long-range dependencies \cite{dosovitskiy2020image,wang2021pyramid,liu2021swin,ding2022davit,chen2022mixformer}, benefiting many computer vision tasks including image SR \cite{liang2021swinir,fang2022hnct,zhang2022elan,choi2023ngram,chen2023DAT}. An essential component in popular transformer-based SR methods is the window self-attention (W-SA) \cite{liu2021swin,liu2022swinv2,liang2021swinir,choi2023ngram}.
By bringing locality into self-attention, the W-SA mechanism not only better utilizes spatial information from input images but also mitigates the computational burden when processing high-resolution images \cite{liu2021swin,liang2021swinir}. However, current transformer-based SR methods often employ W-SA with fixed small window sizes, \eg, $8\times8$ in SwinIR \cite{liang2021swinir}, limiting the receptive field to a single scale and preventing the network from gathering multi-scale information such as local textures and repetitive patterns \cite{kim2016vdsr,lim2017edsr,lai2017lapsrn,hui2019imdn}.
In addition, the quadratic computational complexity of W-SA to the window size also makes the expansion of receptive fields unaffordable in practice. 
% introduce the drawbacks of current methods
To mitigate the computational overhead, previous attempts often reduce channels to support large windows, \eg, channel splitting of group-wise multi-scale self-attention (GMSA) in ELAN~\cite{zhang2022elan} and channel compression of permuted self-attention block (PSA) in SRFormer~\cite{Zhou_2023srformer}. However, these methods not only suffer from the trade-off between spatial and channel information but also remain quadratic complexity to window sizes, limiting the window scaling (max $16\times16$ in ELAN~\cite{zhang2022elan} and $24\times24$ in SRFormer~\cite{Zhou_2023srformer} \vs $64\times64$ and larger in ours).
% Previous methods attempt to utilize multi-scale features and large windows for SR, \eg, group-wise multi-scale self-attention (GMSA) in ELAN \cite{zhang2022elan} and permuted self-attention block (PSA) in SRFormer \cite{Zhou_2023srformer}. However, both GMSA and PSA reduce channels to alleviate the computation of large windows, which not only suffer from the trade-off between spatial and channel information but also remain \textbf{quadratic complexity} to window sizes, limiting the window scaling (max $16\times16$ in ELAN and $24\times24$ in SRFormer \vs $64\times64$ and larger in ours).
% conclude
Therefore, how to effectively aggregate multi-scale features while maintaining computational efficiency remains a critical problem for transformer-based SR approaches. 

\par 
To this end, we develop a general strategy to convert popular transformer-based SR networks to hierarchical transformers for efficient image SR (HiT-SR). Motivated by the success of multi-scale feature aggregation in SR \cite{kim2016vdsr,lai2017lapsrn,hui2019imdn,zhang2022elan}, we first propose to replace the fixed small windows with expanding hierarchical windows in transformer layers, enabling HiT-SR to leverage informative multi-scale features with gradually enlarging receptive fields. To cope with the increasing computational burdens of W-SA in handling large windows, we further design a spatial-channel correlation (SCC) method for efficient aggregation of hierarchical features. Specifically, our SCC consists of a dual feature extraction (DFE) layer to improve feature projection by combining spatial and channel information, a spatial and channel self-correlation (S-SC and C-SC) approach to efficiently exploit hierarchical features with \textbf{linear computational complexity} to window sizes, better supporting window scaling.
In addition, unlike the conventional W-SA that employs hardware inefficient softmax layers \cite{cai2023efficientvit} and time-consuming window shifting operations, our SCC directly uses feature correlation matrices for transformation and hierarchical windows for receptive field expansion, boosting computational efficiency while preserving functionality.

% In addition, unlike the conventional W-SA, our SSC drops time-consuming operations, \eg, hardware inefficient softmax layer \cite{cai2023efficientvit} and window shifting

% our S-SC and C-SC drop the hardware inefficient softmax layer \cite{cai2023efficientvit} and directly use feature correlation matrices for transformation, boosting computational efficiency while preserving functionality. 
% Finally, different head
\par 
Overall, our main contributions are three-fold:
\begin{itemize}
    \item We propose a simple yet effective strategy, \ie, HiT-SR, to convert popular transformer-based SR methods to our hierarchical transformers, boosting SR performance by exploiting multi-scale features and long-range dependencies.
    % leveraging multi-scale features and  for better SR performance.
    \item We design a spatial-channel correlation method to efficiently leverage spatial and channel features with linear computational complexity to window sizes, enabling utilization of large hierarchical windows, \eg, $64\times64$ windows.
    \item We convert SwinIR-Light \cite{liang2021swinir}, SwinIR-NG~\cite{choi2023ngram} and SRFormer-Light~\cite{Zhou_2023srformer} to HiT-SR versions, \ie, HiT-SIR, HiT-SNG, and HiT-SRF, achieving better performance with fewer parameters, FLOPs, and $\sim 7\times$ speed-up (Fig.~\ref{fig:first}).
    % \item We apply our HiT-SR method to improve previous SR transformers SwinIR-Light \cite{liang2021swinir}, SwinIR-NG~\cite{choi2023ngram} and SRFormer-Light~\cite{Zhou_2023srformer}, corresponding to HiT-SIR, HiT-SNG, and HiT-SRF, which achieve state-of-the-art SR performance with fewer parameters, FLOPs, and $\sim 7\times$ speed-up. (Fig.~\ref{fig:first}).
\end{itemize}

% \vspace{-3mm}
\section{Related Work}
% \vspace{-2mm}
\par  
\textbf{Efficient SR.}
%% efficient CNN
Several CNN-based methods are proposed to approach SR in an efficient way. Previous works first explore compact building blocks for SR networks~\cite{lim2017edsr,ahn2018carn,du2022fmen}. Meanwhile, information distillation methods are developed to progressively refine image features for better performance \cite{hui2019imdn,liu2020rfdn}, and LatticeNet designs an economical SR structure based on the lattice filter bank~\cite{luo2020latticenet}. 
Recent methods further improve the efficiency of image SR by utilizing network pruning techniques~\cite{zhang2021srpn,wang2023gassl}. However, most CNN-based SR approaches focus on exploiting local features and struggle to utilize long-range dependencies in image SR.
% To facilitate practical use, several CNN-based methods have been proposed to approach SR in an efficient way. Previous works first explore compact building blocks for SR networks, such as residual blocks \cite{lim2017edsr}, cascading blocks \cite{ahn2018carn}, and high-frequency attention blocks \cite{du2022fmen}. Meanwhile, information distillation methods are developed to progressively refine image features for better performance \cite{hui2019imdn,liu2020rfdn}, and LatticeNet designs an economical SR structure based on the lattice filter bank with butterfly structures \cite{luo2020latticenet}. Recent methods further improve the efficiency of image SR by utilizing network pruning techniques, including structure-regularized pruning \cite{zhang2021srpn} and global aligned structured sparsity learning \cite{wang2023gassl}. However, most CNN-based SR approaches focus on exploiting local features and struggle to utilize long-range dependencies in the image SR task.

\par 
\noindent \textbf{Transformer-based SR.}
%% transformer, add omnisr and srformer, drop date?
The self-attention (SA) mechanism has been widely employed to establish long-range dependencies in both high-level \cite{ramachandran2019stand,dosovitskiy2020image,touvron2021training,zheng2021rethinking,liu2021swin,liu2022swinv2,cao2022swinunet} and low-level vision tasks \cite{liang2021swinir,wang2022uformer,zamir2022restormer,chen2021pre}. For the SR task, SwinIR \cite{liang2021swinir} designs a general image restoration framework based on the shifted window self-attention (SW-SA) \cite{liu2021swin}, and several techniques have been developed to utilize a wider range of features, including N-Gram~\cite{choi2023ngram}, omni self-attention (OSA) \cite{wang2023omnisr}, and permuted self-attention (PSA) \cite{Zhou_2023srformer}.
Although many advances have been made, most existing works neglect the importance of hierarchical features in SR due to the intensive computation required by W-SA on large windows. 
% Inspired by the success of transformers in natural language processing, the self-attention (SA) mechanism has been widely employed to establish long-range dependencies in both high-level \cite{ramachandran2019stand,dosovitskiy2020image,touvron2021training,zheng2021rethinking,liu2021swin,liu2022swinv2,cao2022swinunet} and low-level vision tasks \cite{liang2021swinir,wang2022uformer,zamir2022restormer,chen2021pre}. For the SR task, SwinIR \cite{liang2021swinir} designs a general image restoration framework based on the shifted window self-attention (SW-SA) proposed by Swin transformer \cite{liu2021swin}, achieving promising results in SR. Considering the fixed small windows used in SwinIR, several techniques have been developed to utilize a wider range of spatial information, including N-Gram~\cite{choi2023ngram}, omni self-attention (OSA) \cite{wang2023omnisr}, and permuted self-attention (PSA) \cite{Zhou_2023srformer}.
% Apart from expanding receptive fields in the spatial dimension, the recent dual aggregation transformer (DAT) investigates the utilization of features from spatial and channel domains to enhance the SR performance \cite{chen2023DAT}. Although many advances have been made on transformer-based SR, most existing works neglect the importance of hierarchical features in SR due to the intensive computation required by W-SA on large windows. 

\par 
\noindent \textbf{Hierarchical Feature.}
%% importance of hierarchical feat, and how to do it efficiently
Hierarchical features have been proven effective in boosting SR performance \cite{kim2016vdsr,lim2017edsr,lai2017lapsrn,hui2019imdn,zhang2022elan}, 
but it is generally challenging for transformer-based SR methods to utilize hierarchical features due to the quadratic complexity of W-SA to window sizes. Recent method ELAN designs a group-wise multi-scale SA (GMSA) technique to gather features at different scales \cite{zhang2022elan}. However, GMSA remains quadratic complexity and suffers from information trade-off due to channel splitting. In our approach, a spatial-channel correlation (SCC) method is designed to efficiently utilize hierarchical features with linear complexity to window sizes, while preserving spatial and channel information.
% Hierarchical features have been proven effective in boosting SR performance \cite{kim2016vdsr,lim2017edsr,lai2017lapsrn,hui2019imdn,zhang2022elan}. Previous works first explore multi-scale architectures to take advantage of inter-scale correlation, achieving improvements in both performance and efficiency \cite{kim2016vdsr,lim2017edsr}. Besides, LapSRN proposes a Laplacian pyramid SR network to progressively restore HR images in a coarse-to-fine manner \cite{lai2017lapsrn}, and IMDN also develops a progressive refinement block to step-by-step extract hierarchical features from input LR images \cite{hui2019imdn}. Unlike the aforementioned CNN-based SR approaches, it is more challenging for transformer-based SR methods to utilize hierarchical features due to the quadratic computational complexity of W-SA to window sizes. Recent method ELAN designs a group-wise multi-scale SA (GMSA) technique to aggregate features at different scales \cite{zhang2022elan}. However, the channel-splitting operation in ELAN will result in a trade-off between spatial and channel information, and the computational complexity of GMSA remains quadratic to window sizes. In our approach, a spatial-channel correlation (SCC) method is designed to efficiently utilize hierarchical features with linear computational complexity to window sizes, while preserving spatial and channel information.

\section{Method}
We first introduce the basic methodology of HiT-SR in Sec.~\ref{sec:hit_framework} and then present the block-level and layer-level designs in HiT-SR in Sec.~\ref{sec:block_level} and \ref{sec:layer_level}, respectively.

\subsection{Hierarchical Transformer}\label{sec:hit_framework}
\begin{figure*}[!t]
    \centering
    \includegraphics[width=0.8\linewidth]{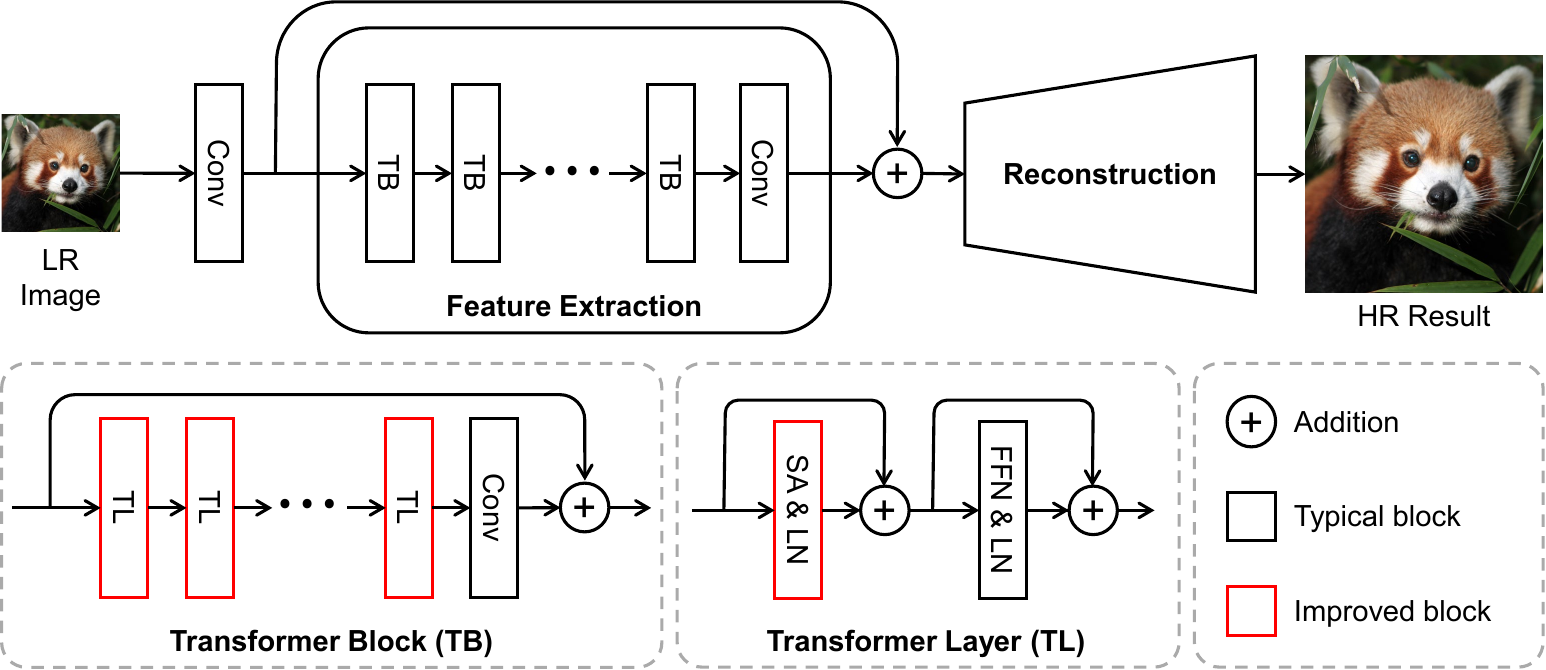}
    % \vspace{-0.8em}
    \caption{Typical framework for transformer-based SR methods, where the block-level and layer-level improvements made by our HiT-SR are colored \textcolor{red}{red}. SA, FFN, and LN indicate self-attention, feed-forward network, and layer normalization, respectively. }
    \label{fig:framework}
    % \vspace{-1.7em}
\end{figure*}

We first review the commonly adopted pipeline of transformer-based SR methods \cite{liang2021swinir,choi2023ngram,Zhou_2023srformer}. As depicted in Fig.~\ref{fig:framework}, popular transformer-based SR framework often consists of convolutional layers to extract shallow features $F_{S} \in \mathbb{R}^{C\times H\times W}$ from LR input images $I_{LR} \in \mathbb{R}^{3\times H\times W}$, a feature extraction module to aggregate deep image features $F_{D} \in \mathbb{R}^{C\times H\times W}$ via transformer blocks (TBs), and a reconstruction module to restore HR images $I_{HR} \in \mathbb{R}^{3\times sH\times sW}$ ($s$ denotes upscaling factor) from shallow and deep features. In the feature extraction module, TBs are generally built with cascaded transformer layers (TLs) followed by convolutional layers, where each TL consists of self-attention (SA), feed-forward network (FFN), and layer normalization (LN). Since the computational complexity of SA is quadratic to input sizes \cite{dosovitskiy2020image}, window partition is often employed in TL to limit SA on local regions, which is known as window self-attention (W-SA) \cite{liu2021swin,liang2021swinir}. Although W-SA alleviates the computational burden, its receptive field is restricted to small local regions, hindering SR networks from utilizing long-range dependencies and multi-scale information.

% TL often partitions inputs to small windows, \eg, $8\times 8$ windows, to perform SA on local regions, which is known as window-based self-attention (W-SA) \cite{liu2021swin,liang2021swinir}.

\par 
To efficiently aggregate hierarchical features, we propose a general strategy to convert the above SR framework to hierarchical transformers. As shown in Fig.~\ref{fig:framework}, our approach mainly contains improvements in two aspects: (i) Instead of using fixed small window sizes for all TLs, we apply hierarchical windows to different TLs in the block level, enabling HiT-SR to establish long-range dependency and gather multi-scale information; (ii) To overcome the computational burden caused by large windows, we replace the W-SA in TLs with a novel spatial-channel correlation (SCC) method, which better supports window scaling with linear computational complexity to window sizes. 
% whose computational complexity is linear to window sizes. 
Based on the above strategies, HiT-SR not only gains better performance by exploiting hierarchical features but also maintains computational efficiency thanks to SCC. 

\subsection{Block-Level Design: Hierarchical Windows}\label{sec:block_level}
\begin{wrapfigure}{r}{0.5\linewidth}
\centering
% \vspace{-2em}
\includegraphics[width=0.8\linewidth]{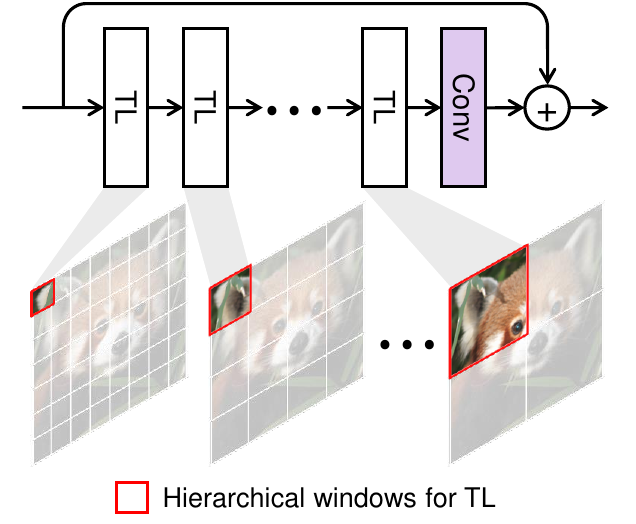} 
% \vspace{-3mm}
\caption{Block-Level design in HiT-SR. Hierarchical windows are applied to different transformer layers (TLs) to aggregate features with expanding receptive fields.}
    \label{fig:hierwin}
    % \vspace{-1.5em}
\end{wrapfigure}

At the block level, we assign hierarchical windows to different TLs, collecting multi-scale features. 
Given a base window size $h_{B}\times w_{B}$, we set the window size $h_i\times w_i$ for the $i$-th TL to
\begin{equation}
    h_i = \alpha_i h_{B},\quad w_i = \alpha_i w_{B},
\end{equation}
where $\alpha_i>0$ is the hierarchical ratio for the $i$-th TL. 
% Then, we divide the input features into non-overlapped windows based on the assigned window sizes for further processing.

\begin{figure*}[!t]
    \centering
    \includegraphics[width=0.9\linewidth]{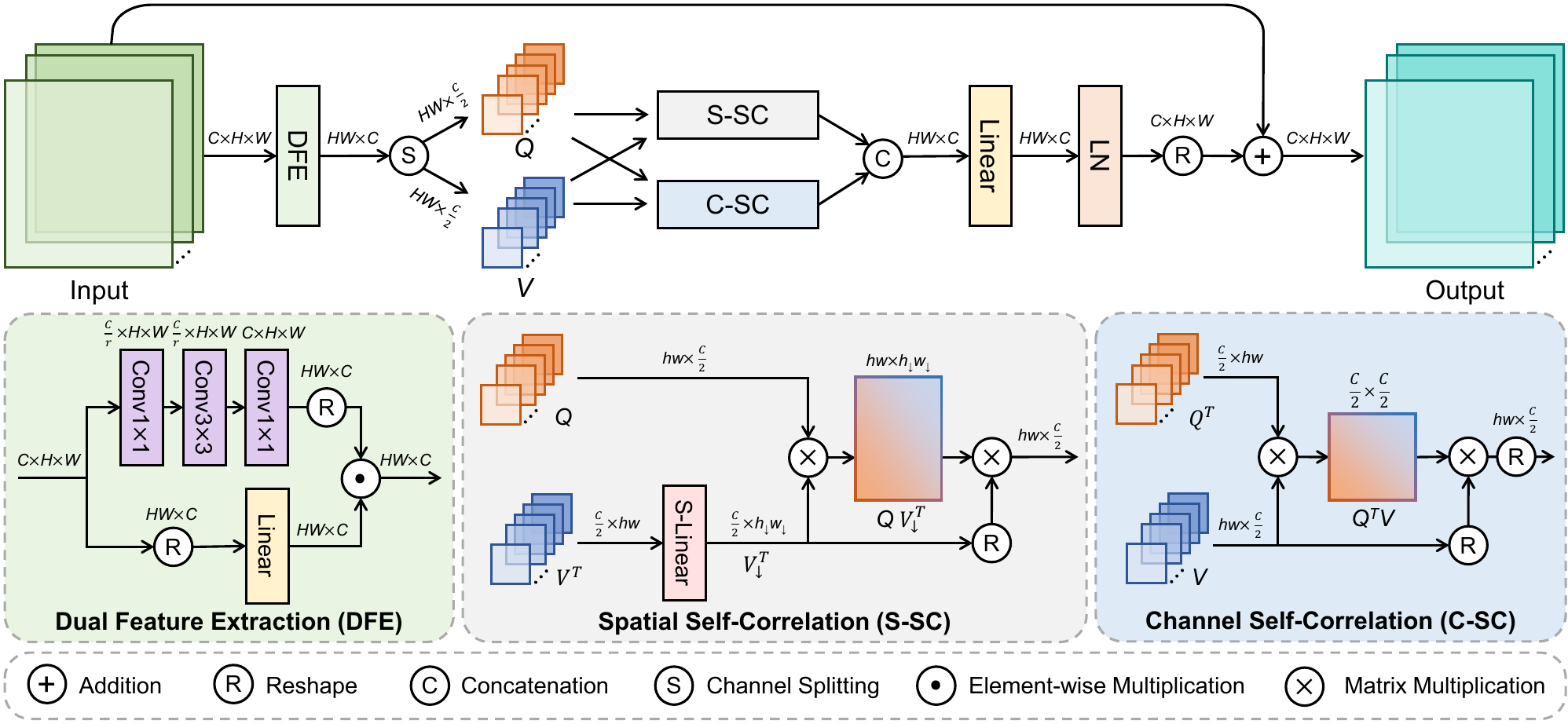}
    % \vspace{-2em}
    \caption{Layer-Level design in HiT-SR composed of dual feature extraction (DFE), spatial and channel self-correlation (S-SC and C-SC). DFE is designed to extract features from spatial and channel domains. S-SC and C-SC are proposed to efficiently aggregate hierarchical information with linear computational complexity to window sizes. }
    % Window partition and reverse processes are omitted for readability.  }
    \label{fig:dualcorr}
    % \vspace{-1.5em}
\end{figure*}
\par 
\noindent \textbf{Expanding Windows.}
% fig, why use expanding, and compare to shifted wins here?
To better aggregate hierarchical features, we arrange the windows with an expanding strategy. As illustrated in Fig.~\ref{fig:hierwin}, we first use small window sizes in the initial layers to gather the most relevant features from local regions, and then gradually expand the window size to utilize the information gained from long-range dependencies. Previous approaches often apply shifting and masking operations on fixed small windows \cite{liang2021swinir,choi2023ngram,Zhou_2023srformer,chen2023DAT} to enlarge receptive fields, but these operations are time-consuming and inefficient in practice. Compared with them, our approach directly utilizes the cascaded TLs to form a hierarchical feature extractor, enabling small to large receptive fields while maintaining overall efficiency. Fig.~\ref{fig:first} shows the better performance of our HiT-SR methods with $\sim 7\times$ faster speeds over the original models.

% As shown in Fig.~\ref{fig:first}, our HiT-SR methods show better performance with $\sim 7\times$ faster speeds compared with the original versions.

\subsection{Layer-Level Design: Spatial-Channel Correlation}\label{sec:layer_level}
At the layer level, we propose spatial-channel correlation (SCC) to efficiently leverage spatial and temporal information from hierarchical inputs. As depicted in Fig.~\ref{fig:dualcorr}, our SCC mainly consists of dual feature extraction (DFE), spatial self-correlation (S-SC), and channel self-correlation (C-SC). Besides, unlike commonly adopted multi-head strategies \cite{
vaswani2017attention,liang2021swinir,chen2023DAT}, different correlation head strategies are applied to S-SC and C-SC to better utilize image features.

\par 
\noindent \textbf{Dual Feature Extraction.}
% linear
Linear layers are often employed for feature projection, which only extracts channel information and neglects to model spatial relations. Instead, we propose DFE with a two-branch design to utilize features from two domains. As shown in Fig.~\ref{fig:dualcorr}, DFE consists of a convolutional branch to exploit spatial information and a linear branch to extract channel features. Given an input feature $X \in \mathbb{R}^{C\times H\times W}$, the output of DFE is computed as
\begin{equation}
\begin{aligned}
&\operatorname{DFE}(X) = X_{ch} \odot X_{sp},\quad \text{with}
\\
X&_{ch} = \operatorname{Linear}(X),\ X_{sp} = \operatorname{Conv}(X),
\end{aligned}
\end{equation}
where $\odot$ denotes element-wise multiplication. The reshaped channel feature $X_{ch} \in \mathbb{R}^{HW\times C}$ and spatial feature $X_{sp} \in \mathbb{R}^{HW\times C}$ are captured by linear and convolutional layers, respectively. In the spatial branch, we use an hourglass structure to stack three convolutional layers with the hidden dimension reduced by ratio $r$ for efficiency. Finally, the spatial and channel features interact with each other by multiplication, yielding the DFE output. 
% Standard SA methods often predict queries, keys, and values for self-attention. 
\par 
Unlike standard SA methods that predict queries, keys, and values by linear projection, we equate keys with values as they both reflect the intrinsic properties of input features, and only estimate queries $Q\in \mathbb{R}^{HW\times \frac{C}{2}}$ and values $V \in \mathbb{R}^{HW\times \frac{C}{2}}$ by splitting the DFE output as displayed in Fig.~\ref{fig:dualcorr},
\begin{equation}
    [Q, V] = \operatorname{DFE}(X),
\end{equation}
which reduces the information redundancy caused by key estimation. Then, we partition queries and keys to non-overlapped windows according to the assigned window size, \eg, $Q_i,\ V_i\in \mathbb{R}^{h_{i}w_{i}\times \frac{C}{2}}$ for the $i$-th TL (the number of windows is omitted for simplicity), and use the partitioned queries and values for the subsequent self-correlation.
% Thus, our approach reduces the information redundancy caused by keys while maintaining SR performance by the subsequent self-correlation on queries and values.

\par 
\noindent \textbf{Spatial Self-Correlation.}
Compared with W-SA, our S-SC aggregates spatial information in an efficient manner. 
% Considering the expanding window sizes in our hierarchical strategy, we first apply a spatial linear layer (denoted as S-Linear and detailed in supplementary material) to adaptively summarize the spatial information of values $V_i$ in different TLs, \ie,
Considering the expanding window sizes in our hierarchical strategy, we first adaptively summarize the spatial information of values $V_i$ in different TLs by applying linear layers on the spatial dimension (denoted as S-Linear and detailed in supplementary material), \ie,
\begin{equation}
    V_{\downarrow,i}^T = \operatorname{S-Linear}_{i}(V_i^T),
\end{equation}
where $V_{\downarrow,i}\in \mathbb{R}^{h_\downarrow w_\downarrow \times \frac{C}{2}}$ denotes the projected values with
\begin{equation}
\left[h_\downarrow, w_\downarrow \right]= \left\{
	\begin{array}{ll}
	\left[h_i, w_i\right], & \text { if } \alpha_i \leq 1, \\
	\left[h_B, w_B\right], & \text { if } \alpha_i > 1.
	\end{array}\right.
\end{equation}
% \begin{equation}
%     [h_\downarrow, w_\downarrow] = \left\{
% 	\begin{array}{ll}
% 	[h_i, w_i], & \text { if } \alpha_i \leq 1, \\
% 	[h_B, w_B], & \text { if } \alpha_i > 1.
% 	\end{array}\right.
% \end{equation}
% Compared with W-SA, our S-SC aggregates spatial information in an efficient manner. Considering the large window sizes in our hierarchical strategy, we first project the partitioned values $V_i$ to the base window size via a spatial linear layer as illustrated in Fig.~\ref{fig:dualcorr}, \ie,
% \begin{equation}
%     V_{\downarrow,i}^T = \operatorname{Linear}_{i}(V_i^T),
% \end{equation}
% where $V_{\downarrow,i}\in \mathbb{R}^{h_B w_B\times \frac{C}{2}}$ represents the projected values, and 
% \begin{equation}
% \operatorname{Linear}_{i}(\cdot): \left\{
% 	\begin{array}{ll}
% 	\mathbb{R}^{\frac{C}{2}\times h_i w_i} \rightarrow \mathbb{R}^{\frac{C}{2}\times h_i w_i}, & \text { if } \alpha_i \leq 1, \\
% 	\mathbb{R}^{\frac{C}{2}\times h_B w_B \times \alpha_i^2} \rightarrow \mathbb{R}^{\frac{C}{2}\times h_B w_B}, & \text { if } \alpha_i > 1.
% 	\end{array}\right.
% \end{equation}
% $\operatorname{Linear}_{i}(\cdot):\mathbb{R}^{\frac{C}{2}\times h_B w_B \times \alpha_i^2} \rightarrow \mathbb{R}^{\frac{C}{2}\times h_B w_B}$ functions on the spatial domain to adaptively summarize the spatial information in different TLs. 
Thus, our HiT-SR is able to summarize high-level information from large windows, \ie, $\alpha_i> 1$, and simultaneously preserve fine-grained features with small windows, \ie, $\alpha_i\leq 1$. Afterward, we compute the S-SC based on $Q_i$ and $V_{\downarrow,i}$ as
\begin{equation}\label{eq:ssc}
    \operatorname{S-SC}(Q_i, V_{\downarrow,i}) = \left(\frac{Q_i V_{\downarrow,i}^T}{D} + B\right)\cdot V_{\downarrow,i},
\end{equation}
where $B$ denotes the relative position encoding \cite{wang2023crossformer++}, and the constant denominator $D=\frac{C}{2}$ is used for normalization. Compared with the standard W-SA, our S-SC shows advantages in efficiency and complexity: (i) We utilize correlation maps instead of attention maps to aggregate information, dropping the hardware inefficient operation softmax to improve inference speeds \cite{cai2023efficientvit};
(ii) Our S-SC supports large windows with linear computational complexity to window sizes. Supposing the input contains $N$ windows with each window in the $\mathbb{R}^{hw\times C}$ space, the numbers of mult-add operations required for W-SA and S-SC are
\begin{equation}\label{eq:ssc_complex}
\begin{aligned}
&\operatorname{Mult-Add}(\operatorname{W-SA})= 2NC(hw)^2,
\\
&\operatorname{Mult-Add}(\operatorname{S-SC})= 2NCh_\downarrow w_\downarrow hw,
\end{aligned}
\end{equation}
where the former is quadratic to window sizes $hw$.
Since $h_\downarrow w_\downarrow$ is upper bounded by the fixed base window size $h_B w_B$, the computational complexity of our S-SC is linear to the window size, benefiting window scaling-up.

% (ii) Our S-SC supports large windows with better computational efficiency. Supposing the input contains $N$ windows with each window in the $\mathbb{R}^{hw\times C}$ space, the computational complexity of W-SA is 
% \begin{equation}
%     \Omega(\operatorname{W-SA}) = 2 N C (h w)^2 m + N (h w)^2 s. 
% \end{equation}
% $m$ for time of one mult-add. $s$ for time of one exp.
% The first term stands for matrix multiplication, and the second term is caused by the softmax layer, resulting in \textit{quadratic} complexity to window sizes. By contrast, the complexity of our S-SC is
% \begin{equation}\label{eq:ssc_complex}
%     \Omega(\operatorname{S-SC}) = 2 N C h_\downarrow w_\downarrow h w.
% \end{equation}
% Since $h_\downarrow\times w_\downarrow$ is upper bounded by the fixed base window size $h_B\times w_B$, the computational complexity of our S-SC is \textit{linear} to the window size, benefiting window scaling-up.

\begin{table}[t]
\scriptsize
\setlength\tabcolsep{5pt}
\caption{Complexity of different layer types. $C, N, h, w$ correspond to channel dimension, window number, height, and width. $m=\operatorname{max}(C,h_\downarrow w_\downarrow)$ is upper bounded.}
\label{table:complexity}
% \vspace{-2.7em}
\begin{center}
\begin{tabular}{lcc}
\toprule[0.15em]
\rowcolor{color3}
Layer Type   & Complexity per layer & Summary  \\ 
\midrule[0.15em]
Global Self-Attention & $O(C\cdot N^2\cdot h^2\cdot w^2)$ & \text{Quadratic to image size}
\\
Window Self-Attention & $O(C\cdot N\cdot h^2\cdot w^2)$ & \text{Quadratic to window size}
\\
\rowcolor{color3} \textbf{Spatial-Channel Correlation (Ours)} & $O(C\cdot N\cdot m\cdot h\cdot w)$ & \text{\textbf{Linear} to window size}
\\
\bottomrule[0.15em]
\end{tabular}
\end{center}
% \vspace{-3.3em}
\end{table}

\par 
\noindent \textbf{Channel Self-Correlation.}
% complexity
Apart from spatial information, we further design C-SC to gather features from the channel domain, as depicted in Fig.~\ref{fig:dualcorr}. Given the partitioned queries and values in the $i$-th TL, the output of C-SC is
\begin{equation}\label{eq:csc}
    \operatorname{C-SC}(Q_i, V_i) = \frac{Q_i^T V_i}{D_i} \cdot V^T_i,
\end{equation}
where the denominator $D_i = h_i w_i$. Compared with the prevalent transposed attention for channel aggregation \cite{ali2021xcit,zamir2022restormer,chen2023DAT}, our C-SC benefits from hierarchical windows and utilizes rich multi-scale information to boost SR performance. For computational complexity, the mult-add operations needed for C-SC is 
\begin{equation}\label{eq:csc_complex}
    \operatorname{Mult-Add}(\operatorname{C-SC}) = 2N C^2 hw
\end{equation}
under inputs in the $\mathbb{R}^{N\times hw \times C}$ space. Combining Eq.~\eqref{eq:ssc_complex} and \eqref{eq:csc_complex}, the complexity of our spatial-channel correlation maintains \textbf{linear} to window sizes as noted in Tab.~\ref{table:complexity}, enabling scalable windows to make full use of hierarchical information.

\par 
\noindent \textbf{Different Correlation Head.} Multi-head strategy is commonly employed in SA to gather information from different representation subspaces \cite{vaswani2017attention}, and it has exhibited promising performance when dealing with spatial information \cite{liang2021swinir,chen2023DAT}. However, when processing channel information, the multi-head strategy instead restricts the receptive field of channel information aggregation, \ie, each channel can only interact with a limited set of other channels, leading to sub-optimal performance. To address this, we propose to apply the standard multi-head strategy to S-SC but use a single-head strategy in C-SC, enabling full channel interaction. Therefore, the S-SC can utilize information from different channel subspaces by the multi-head strategy, and the C-SC can exploit information from different spatial subspaces via hierarchical windows.

\begin{table*}[!ht]
\tiny
\caption{Quantitative comparison with state-of-the-art SR methods. The output size is set to $720 \times 1280$ for all scales to compute FLOPs. Best results are colored \textcolor{red}{red}.
% The best and second-best SR results are colored \textcolor{red}{red} and \textcolor{blue}{blue}, respectively.
% \textcolor{red}{(modify the complexity of HiT-SNG)}
}
\label{table:quantires}
% \vspace{-3.5em}
\begin{center}
\setlength\tabcolsep{1.7pt}
\begin{tabular}{l|c|cc|ccccccccccccccc} 
\toprule[0.15em]
	\rowcolor{color3}  & & \multicolumn{2}{c|}{Complexity} & \multicolumn{2}{c}{Set5} & \multicolumn{2}{c}{Set14} & \multicolumn{2}{c}{B100} & \multicolumn{2}{c}{Urban100} & \multicolumn{2}{c}{Manga109}\\

	\rowcolor{color3} \multirow{-2}{*}{Method} &  \multirow{-2}{*}{Scale} & \#Para. & FLOPs & PSNR & SSIM & PSNR & SSIM & PSNR & SSIM & PSNR & SSIM & PSNR & SSIM
\\
\midrule[0.15em]
EDSR-B \cite{lim2017edsr}  & $\times$2 & 1370K & 316.3G &
37.99 & 0.9604 & 33.57 & 0.9175 & 32.16 & 0.8994 & 31.98 & 0.9272 & 38.54 & 0.9769\\
CARN \cite{ahn2018carn}  & $\times$2 & 1592K & 222.8G &
37.76 & 0.9590 & 33.52 & 0.9166 & 32.09 & 0.8978 & 31.92 & 0.9256 & 38.36 & 0.9765\\
IMDN \cite{hui2019imdn} & $\times$2 & 694K & 158.8G &
38.00 & 0.9605 & 33.63 & 0.9177 & 32.19 & 0.8996 & 32.17 & 0.9283 & 38.88 & 0.9774\\
LatticeNet \cite{luo2020latticenet} & $\times$2 & 756K & 169.5G &
38.06  & 0.9607 & 33.70 & 0.9187 & 32.20 & 0.8999 & 32.25 & 0.9288 & 38.94 & 0.9774\\
RFDN-L \cite{liu2020rfdn}  & $\times$2 & 626K & 145.8G &
38.08 & 0.9606 & 33.67 & 0.9190 & 32.18 & 0.8996 & 32.24 & 0.9290 & 38.95 & 0.9773\\
SRPN-Lite \cite{zhang2021srpn}  & $\times$2 & 609K & 139.9G &
38.10 & 0.9608 & 33.70 & 0.9189 & 32.25 & 0.9005 & 32.26 & 0.9294 & - & -\\
FMEN \cite{du2022fmen}  & $\times$2 & 748K & 172.0G &
38.10 & 0.9609 & 33.75 & 0.9192 & 32.26 & 0.9007 & 32.41 & 0.9311 & 38.95 & 0.9778\\
GASSL-B \cite{wang2023gassl}  & $\times$2 & 689K & 158.2G &
38.08 & 0.9607 & 33.75 & 0.9194 & 32.24 & 0.9005 & 32.29 & 0.9298 & 38.92 & 0.9777\\
SwinIR-L \cite{liang2021swinir}  & $\times$2 & 910K & 244.4G &
38.14 & 0.9611 & 33.86 & 0.9206 & 32.31 & 0.9012 & 32.76 & 0.9340 & 39.12 & 0.9783\\
HNCT \cite{fang2022hnct}  & $\times$2 & 357K & 82.4G &
38.08 & 0.9608 & 33.65 & 0.9182 & 32.22 & 0.9001 & 32.22 & 0.9294 & 38.87 & 0.9774\\
ELAN-L \cite{zhang2022elan}  & $\times$2 & 621K & 201.3G &
\textcolor{black}{38.17} & 0.9611 & 33.94 & 0.9207 & 32.30 & 0.9012 & 32.76 & 0.9340 & 39.11 & 0.9782\\
Omni-SR \cite{wang2023omnisr}  & $\times$2 & 772K & 194.5G & 38.22 & 0.9613 & 33.98 & 0.9210 & 32.36 & 0.9020 & 33.05 & 0.9363 & 39.28 & 0.9784 \\
SwinIR-NG \cite{choi2023ngram} & $\times$2 & 1181K & 274.1G &
\textcolor{black}{38.17} & \textcolor{black}{0.9612} & 33.94 & 0.9205 & 32.31 & 0.9013 & 32.78 & 0.9340 & 39.20 & 0.9781\\
SRFormer-L \cite{Zhou_2023srformer}  & $\times$2 & 853K & 236.2G & 38.23 & 0.9613 & 33.94 & 0.9209 & 32.36 & 0.9019 & 32.91 & 0.9353 & 39.28 & 0.9785 \\
\rowcolor{color3} \textbf{HiT-SIR (Ours)}  & $\times$2 & 772K & 209.9G & 
38.22 & 0.9613 & 33.91 & 0.9213 & 32.35 & 0.9019 & 33.02 & 0.9365 & 39.38  & 0.9782 \\
\rowcolor{color3} \textbf{HiT-SNG (Ours)}  & $\times$2 & \textcolor{black}{1013K} & \textcolor{black}{213.9G}  & 
 38.21 & 0.9612 & 34.00  & \textcolor{red}{0.9217} & 32.35 & 0.9020 & 33.01 & 0.9360 & 39.32 & 0.9782 \\
 \rowcolor{color3} \textbf{HiT-SRF (Ours)}  & $\times$2 & 847K & 226.5G 
 & \textcolor{red}{38.26} & \textcolor{red}{0.9615} & \textcolor{red}{34.01} & 0.9214 & \textcolor{red}{32.37} & \textcolor{red}{0.9023} & \textcolor{red}{33.13} & \textcolor{red}{0.9372} & \textcolor{red}{39.47} & \textcolor{red}{0.9787} \\

%% x3 scale ======================================
\midrule
EDSR-B \cite{lim2017edsr}  & $\times$3 & 1555K & 160.2G &
34.37 & 0.9270 & 30.28  & 0.8417 & 29.09 & 0.8052 & 28.15 & 0.8527 & 33.45 & 0.9439\\
CARN \cite{ahn2018carn}  & $\times$3 & 1592K & 118.8G &
34.29 & 0.9255 & 30.29 & 0.8407 & 29.06 & 0.8034 & 28.06 & 0.8493 & 33.50 & 0.9440\\
IMDN \cite{hui2019imdn}  & $\times$3 & 703K & 71.5G &
34.36 & 0.9270 & 30.32 & 0.8417 & 29.09 & 0.8046 & 28.17 & 0.8519 & 33.61 & 0.9445\\
LatticeNet \cite{luo2020latticenet}  & $\times$3 & 765K & 76.3G &
34.40  & 0.9272 & 30.32 & 0.8416 & 29.10 & 0.8049 & 28.19 & 0.8513 & 33.63 & 0.9442\\
RFDN-L \cite{liu2020rfdn}  & $\times$3 & 633K & 65.6G &
34.47 & 0.9280 & 30.35 & 0.8421 & 29.11 & 0.8053 & 28.32 & 0.8547 & 33.78 & 0.9458\\
SRPN-Lite \cite{zhang2021srpn}  & $\times$3 & 615K & 62.7G &
34.47 & 0.9276 & 30.38 & 0.8425 & 29.16 & 0.8061 & 28.22 & 0.8534 & - & -\\
FMEN \cite{du2022fmen}  & $\times$3 & 757K & 77.2G &
34.45 & 0.9275 & 30.40 & 0.8435 & 29.17 & 0.8063 & 28.33 & 0.8562 & 33.86 & 0.9462\\
GASSL-B \cite{wang2023gassl} & $\times$3 & 691K & 70.4G &
34.47 & 0.9278 & 30.39 & 0.8430 & 29.15 & 0.8063 & 28.27 & 0.8546 & 33.77 & 0.9455\\
SwinIR-L \cite{liang2021swinir}  & $\times$3 & 918K & 110.8G &
34.62 & 0.9289 & 30.54 & 0.8463 & 29.20 & 0.8082 & 28.66 & 0.8624 & 33.98 & 0.9478\\
HNCT \cite{fang2022hnct}  & $\times$3 & 363K & 37.8G &
34.47 & 0.9275 & 30.44 & 0.8439 & 29.15 & 0.8067 & 28.28 & 0.8557 & 33.81 & 0.9459\\
ELAN-L \cite{zhang2022elan}  & $\times$3 & 629K & 89.5G &
34.61 & 0.9288 & 30.55 & 0.8463 & 29.21 & 0.8081 & 28.69 & 0.8624 & 34.00 & 0.9478\\
Omni-SR \cite{wang2023omnisr}  & $\times$3 & 780K  & 88.4G & 34.70 & 0.9294 & 30.57 & 0.8469 & 29.28 & 0.8094 & 28.84 & 0.8656 & 34.22 &  0.9487 \\
SwinIR-NG \cite{choi2023ngram}  & $\times$3 & 1190K & 114.1G &
34.64 & 0.9293 & \textcolor{black}{30.58} & 0.8471 & \textcolor{black}{29.24} & \textcolor{black}{0.8090} & 28.75 & 0.8639 & 34.22 & 0.9488\\
SRFormer-L \cite{Zhou_2023srformer}  & $\times$3 & 861K & 104.8G & 34.67 & 0.9296 & 30.57 & 0.8469 & 29.26 & 0.8099 & 28.81 & 0.8655 & 34.19 & 0.9489  \\
\rowcolor{color3} \textbf{HiT-SIR (Ours)}  & $\times$3 & 780K & 94.2G &  34.72 & 0.9298 & \textcolor{red}{30.62} & 0.8474 & 29.27 & 0.8101 & 28.93 & 0.8673 & 34.40 & 0.9496  \\
\rowcolor{color3} \textbf{HiT-SNG (Ours)}  & $\times$3 & \textcolor{black}{1021K} & \textcolor{black}{99.5G} & 34.74 & 0.9297 & \textcolor{red}{30.62} & 0.8474 & 29.26 & 0.8100 & 28.91 & 0.8671 & 34.38 & 0.9495  \\
\rowcolor{color3} \textbf{HiT-SRF (Ours)}  & $\times$3 & 855K & 101.6G & \textcolor{red}{34.75} & \textcolor{red}{0.9300} & 30.61 & \textcolor{red}{0.8475} & \textcolor{red}{29.29} & \textcolor{red}{0.8106} & \textcolor{red}{28.99} & \textcolor{red}{0.8687} & \textcolor{red}{34.53} & \textcolor{red}{0.9502}  \\
 
%% x4 scale ======================================
\midrule
EDSR-B \cite{lim2017edsr}  & $\times$4 & 1518K & 114.0G &
32.09 & 0.8938 & 28.58 & 0.7813 & 27.57 & 0.7357 & 26.04 & 0.7849 & 30.35 & 0.9067\\
CARN \cite{ahn2018carn}  & $\times$4 & 1592K & 90.9G &
32.13 & 0.8937 & 28.60 & 0.7806 & 27.58 & 0.7349 & 26.07 & 0.7837 & 30.47 & 0.9084\\
IMDN \cite{hui2019imdn}  & $\times$4 & 715K & 40.9G &
32.21 & 0.8948 & 28.58 & 0.7811 & 27.56 & 0.7353 & 26.04 & 0.7838 & 30.45 & 0.9075\\
LatticeNet \cite{luo2020latticenet}  & $\times$4 & 777K & 43.6G &
32.18  & 0.8943 & 28.61 & 0.7812 & 27.57 & 0.7355 & 26.14 & 0.7844 & 30.54 & 0.9075\\
RFDN-L \cite{liu2020rfdn}  & $\times$4 & 643K & 37.4G &
32.28 & 0.8957 & 28.61 & 0.7818 & 27.58 & 0.7363 & 26.20 & 0.7883 & 30.61 & 0.9096\\
SRPN-Lite \cite{zhang2021srpn} & $\times$4 & 623K & 35.8G &
32.24 & 0.8958 & 28.69 & 0.7836 & 27.63 & 0.7373 & 26.16 & 0.7875 & - & -\\
FMEN \cite{du2022fmen}  & $\times$4 & 769K & 44.2G &
32.24 & 0.8955 & 28.70 & 0.7839 & 27.63 & 0.7379 & 26.28 & 0.7908 & 30.70 & 0.9107\\
GASSL-B \cite{wang2023gassl} & $\times$4 & 694K & 39.9G &
32.17 & 0.8950 & 28.66 & 0.7835 & 27.62 & 0.7377 & 36.16 & 0.7888 & 30.70 & 0.9100\\
SwinIR-L \cite{liang2021swinir}  & $\times$4 & 930K & 63.6G &
32.44 & 0.8976 & 28.77 & 0.7858 & 27.69 & 0.7406 & 26.47 & 0.7980 & 30.92 & 0.9151\\
HNCT \cite{fang2022hnct}  & $\times$4 & 373K & 22.0G &
32.31 & 0.8957 & 28.71 & 0.7834 & 27.63 & 0.7381 & 26.20 & 0.7896 & 30.70 & 0.9112\\
ELAN-L \cite{zhang2022elan}  & $\times$4 & 640K & 53.7G &
32.43 & 0.8975 & 28.78 & 0.7858 & 27.69 & 0.7406 & 26.54 & 0.7982 & 30.92 & 0.9150\\
Omni-SR \cite{wang2023omnisr}  & $\times$4 & 792K & 50.9G & 32.49 & 0.8988 & 28.78 & 0.7859 & 27.71 & 0.7415 & 26.64 & 0.8018 & 31.02 & 0.9151  \\
SwinIR-NG \cite{choi2023ngram}  & $\times$4 & 1201K & 64.4G &
32.44 & 0.8980 & \textcolor{black}{28.83} & \textcolor{black}{0.7870} & \textcolor{black}{27.73} & \textcolor{black}{0.7418} & 26.61 & 0.8010 & 31.09 & 0.9161\\
SRFormer-L \cite{Zhou_2023srformer}  & $\times$4 & 873K & 62.8G & 32.51 & 0.8988 & 28.82 & 0.7872 & 27.73 & 0.7422 & 26.67 & 0.8032 & 31.17 & 0.9165  \\
\rowcolor{color3} \textbf{HiT-SIR (Ours)} & $\times$4 & 792K & 53.8G &  32.51 & \textcolor{black}{0.8991} & 28.84 & 0.7873 & 27.73 & 0.7424 & 26.71 & 0.8045 & 31.23 & \textcolor{red}{0.9176}  \\
\rowcolor{color3} \textbf{HiT-SNG (Ours)}  & $\times$4 & \textcolor{black}{1032K} & \textcolor{black}{57.7G} & \textcolor{red}{32.55} & 0.8991 & 28.83 & 0.7873 & \textcolor{black}{27.74} & 0.7426 & 26.75 & 0.8053 & 31.24 & \textcolor{red}{0.9176}  \\
 \rowcolor{color3} \textbf{HiT-SRF (Ours)} & $\times$4 & 866K & 58.0G & \textcolor{red}{32.55} & \textcolor{red}{0.8999} & \textcolor{red}{28.87} & \textcolor{red}{0.7880} & \textcolor{red}{27.75} & \textcolor{red}{0.7432} & \textcolor{red}{26.80} & \textcolor{red}{0.8069} & \textcolor{red}{31.26} & \textcolor{black}{0.9171}  \\
\bottomrule[0.15em]
\end{tabular}
\end{center}
% \vspace{-5.2em}
\end{table*}

\section{Experiments and Analysis}
\subsection{Experimental Settings}\label{sec:exp_setting}
\noindent \textbf{Implementation Details.}
%% SwinIR + SwinNG
We apply our HiT-SR strategy to the popular SR method SwinIR-Light \cite{liang2021swinir} and the recent state-of-the-art SR approaches SwinIR-NG \cite{choi2023ngram} and SRFormer-Light \cite{Zhou_2023srformer}, corresponding to HiT-SIR, HiT-SNG, and HiT-SRF in this paper. To fairly verify the effectiveness and adaptivity of our method, we convert each method to the HiT-SR version with minimal changes and apply the same hyperparameter settings for all SR transformers.
% do not tune hyperparameters for different SR transformers.
Specifically, we follow the original settings of SwinIR-Light \cite{liang2021swinir} and set the TB number, TL number, channel number, and head number of all HiT-SR improved models to 4, 6, 60, and 6, respectively. The base window size $h_{B}\times w_{B}$ are set to the widely adopted value, \ie, $8\times8$, and we set the hierarchical ratios $[0.5, 1, 2, 4, 6, 8]$ for the 6 TLs in each TB. 
% Note that no additional hyperparameter tuning is required
% \begin{itemize}
%     \item \textbf{HiT-SIR.} We mainly follow the original settings of SwinIR-Light (the TB number, TL number, channel number, and head number are set to 4, 6, 60, and 6, respectively) and set the base window size $h_{B}\times w_{B}$ to the original window size of SwinIR-Light, \ie, $8\times8$. The hierarchical ratios are set to $[\frac{1}{2}, 1, 2, 4, 6, 8]$ for the 6 TLs in each TB. 
%     % We further design a more efficient version of HiT-SIR, denoted as HiT-SIR-S, with the same parameter settings except that the channel number is reduced to 48.
%     \item \textbf{HiT-SNG.} \textcolor{black}{Similarly, we follow SwinIR-NG to set the TB number, TL number, channel number, and head number to 4, 6, 60, and 6, respectively. The base window size $h_{B}\times w_{B}$ are set to the original window size of SwinIR-NG, \ie, $8\times8$, and we set the hierarchical ratios $[\frac{1}{2}, 1, 2, 4, 6, 8]$ for the 6 TLs in each TB.   } 
%     \item \textbf{HiT-SRF.} 
% \end{itemize}
\par 
We apply the same training strategy to HiT-SIR, HiT-SNG, and HiT-SRF. All the models are implemented based on PyTorch \cite{paszke2019pytorch} and trained with patch size $64\times64$ and batch size $64$ for 500K iterations. $L_1$ loss and Adam optimizer \cite{kingma2014adam} ($\beta_1=0.9$ and $\beta_2=0.99$) are employed for model optimization. We set the initial learning rate as $5\times10^{-4}$, and half it at [250K,400K,450K,475K] iterations. We also randomly utilize 90°, 180°, and 270° rotations and horizontal flips for data augmentation during model training.

\par 
\noindent \textbf{Data and Evaluation.} Following previous practice \cite{liang2021swinir,choi2023ngram,Zhou_2023srformer}, we employ the popular DIV2K \cite{agustsson2017ntire_div2k} dataset for training and the five classical benchmark datasets: Set5 \cite{bevilacqua2012low_set5}, Set14 \cite{zeyde2012single_set14}, B100 \cite{martin2001database_BSD100}, Urban100 \cite{huang2015single_urban100}, and Manga109 \cite{matsui2017sketch_manga109}, for evaluation. The LR images are obtained from their HR counterparts by bicubic degradation. We conduct comparisons under three upscaling factors: $\times2$, $\times3$, and $\times4$, and assess the SR performance by the two commonly used metrics PSNR and SSIM \cite{wang2004ssim}, which are computed on the Y channel in the YCbCr space.

\def\scale{0.4}
\begin{figure}[!t]
% {r}{0.5\linewidth}
    \centering
%% for performance ==========================
% for params
       \begin{tikzpicture}[scale=\scale]
		\begin{axis}
			[
             height=0.618\linewidth,
             width=\linewidth,
             ymin=31.8, ymax=33.2,
              ylabel={\textbf{PSNR on Urban100 ($\times 2$)}},
              xlabel={Training iterations},
              line width=1.2pt,
			 axis y line=left,
			 axis x line=bottom,
              every axis/.append style={font=\large},
              grid,
              legend pos=south east,
    grid style={color=white, line width=0.5pt},
              axis background/.style={fill=gray!10}, % 设置背景颜色
			 enlarge x limits=0.1, ] 
			\addplot+ [mark=none] table [col sep=comma, x=Step, y=Value] {data/main/SwinIR-DF2K-U100.csv};
                \addlegendentry{SwinIR-Light}
                \addplot+ [mark=none] table [col sep=comma, x=Step, y=Value] {data/main/HiT-SIR-DF2K-U100.csv};
                \addlegendentry{HiT-SIR}

		\end{axis} 
	\end{tikzpicture} 
%% for performance ==========================
% for params
       \begin{tikzpicture}[scale=\scale]
		\begin{axis}
			[
             height=0.618\linewidth,
             width=\linewidth,
             ymin=38.3, ymax=39.7,
              ylabel={\textbf{PSNR on Manga109 ($\times 2$)}},
              xlabel={Training iterations},
              line width=1.2pt,
			 axis y line=left,
			 axis x line=bottom,
              every axis/.append style={font=\large},
              grid,
              legend pos=south east,
    grid style={color=white, line width=0.5pt},
              axis background/.style={fill=gray!10}, % 设置背景颜色
			 enlarge x limits=0.1, ] 
			\addplot+ [mark=none] table [col sep=comma, x=Step, y=Value] {data/main/SwinIR-DF2K-M109.csv};
                \addlegendentry{SwinIR-Light}
                \addplot+ [mark=none] table [col sep=comma, x=Step, y=Value] {data/main/HiT-SIR-DF2K-M109.csv};
                \addlegendentry{HiT-SIR}
                
		\end{axis} 
	\end{tikzpicture} 
    % \vspace{-1em}
    \caption{Convergence comparison of SwinIR-Light \cite{liang2021swinir} and our improved version HiT-SIR on Urban100 ($\times 2$) and Manga109 ($\times 2$) datasets under the same training settings. }
    \label{fig:triancurve}
    % \vspace{-0.8em}
\end{figure}
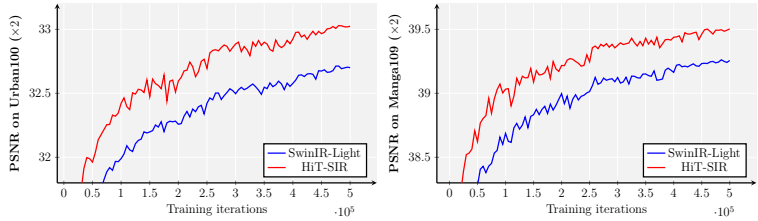

\begin{table*}[!t]
\scriptsize
\caption{Comparisons between state-of-the-art efficient SR transformers and our HiT-SR approaches. Complexity metrics are measured under $\times2$ upscaling on an A100 GPU with the output image size set to $720\times1280$. Improvements are highlighted in \textcolor{red}{red}.
}
\label{table:improvement}
% \vspace{-2.5em}
\begin{center}
\setlength\tabcolsep{3pt}
\begin{tabular}{l|ccc|cc} 
\toprule[0.15em]
	\rowcolor{color3}  & \multicolumn{3}{c|}{Computational complexity} & \multicolumn{2}{c}{\textcolor{black}{Urban100} ($\times2$)} \\
	\rowcolor{color3} \multirow{-2}{*}{Method}  & \#Para. (K) & FLOPs (G) & Runtime (ms) & PSNR & SSIM 
\\
\midrule[0.15em]
SwinIR-Light \cite{liang2021swinir} & 910 & 244.4 & 1899 & 32.76 & 0.9340
\\
\rowcolor{color3} \textbf{HiT-SIR (Ours)} & 772 \textcolor{red}{(138$\downarrow$)} & 209.9 \textcolor{red}{(34.5$\downarrow$)} & 249 \textcolor{red}{(7.6$\times$)} & 33.02 \textcolor{red}{(0.26$\uparrow$)} & 0.9365 \textcolor{red}{(0.0025$\uparrow$)}
\\
\midrule
SwinIR-NG \cite{choi2023ngram} & 1181 & 274.1 & 1920 & 32.78 & 0.9340
\\
\rowcolor{color3} \textbf{HiT-SNG (Ours)} & 1013 \textcolor{red}{(168$\downarrow$)} & 213.9 \textcolor{red}{(60.2$\downarrow$)} & 279 \textcolor{red}{(6.9$\times$)} & 33.01 \textcolor{red}{(0.23$\uparrow$)} & 0.9360 \textcolor{red}{(0.0020$\uparrow$)}
\\
\midrule
SRFormer-Light \cite{Zhou_2023srformer} & 853 & 236.2 & 2593 & 32.91 & 0.9353
\\
\rowcolor{color3} \textbf{HiT-SRF (Ours)} & 847 \textcolor{red}{(6$\downarrow$)} & 226.5 \textcolor{red}{(9.7$\downarrow$)} & 331 \textcolor{red}{(7.8$\times$)} & 33.13 \textcolor{red}{(0.22$\uparrow$)} & 0.9372 \textcolor{red}{(0.0019$\uparrow$)}
\\
\bottomrule[0.15em]
\end{tabular}
\end{center}
% \vspace{-3.5em}
\end{table*}
\subsection{Benchmarking}
% compare setting, quantitative, qualitative, improvements compared with original structure fig 1
We evaluate the proposed HiT-SR by comparing our HiT-SIR, HiT-SNG, and HiT-SRF with existing state-of-the-art efficient SR approaches, including CNN-based algorithms EDSR-B \cite{lim2017edsr}, CARN \cite{ahn2018carn}, IMDN \cite{hui2019imdn}, LatticeNet \cite{luo2020latticenet}, RFDN-L \cite{liu2020rfdn}, SRPN-Lite \cite{zhang2021srpn}, FMEN \cite{du2022fmen}, and GASSL-B \cite{wang2023gassl}, as well as transformer-based methods SwinIR-Light \cite{liang2021swinir}, HCNT \cite{fang2022hnct}, ELAN-Light \cite{zhang2022elan}, Omni-SR \cite{wang2023omnisr}, SwinIR-NG \cite{choi2023ngram}, and SRFormer-Light \cite{Zhou_2023srformer}.

\begin{figure*}[!th]
\tiny
\centering
\scalebox{1}{
\begin{tabular}{cccc}
% % one row
\hspace{-0.2cm}
\begin{adjustbox}{valign=t}
\begin{tabular}{c}
\includegraphics[width=0.219\textwidth]{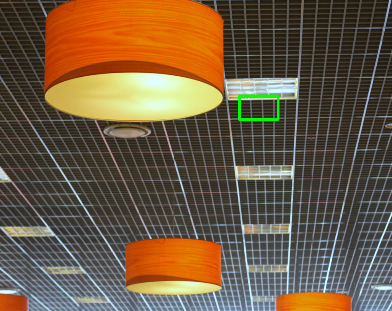}
\\
Urban100: img\_044 
\end{tabular}
\end{adjustbox}
\hspace{-0.23cm}
\begin{adjustbox}{valign=t}
\begin{tabular}{cccccc}
\includegraphics[width=0.149\textwidth]{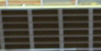} \hspace{-1.5mm} &
\includegraphics[width=0.149\textwidth]{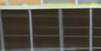} \hspace{-1.5mm} &
\includegraphics[width=0.149\textwidth]{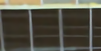} \hspace{-1.5mm} &
\includegraphics[width=0.149\textwidth]{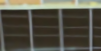} \hspace{-1.5mm} &
\includegraphics[width=0.149\textwidth]{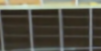} \hspace{-1.5mm} 
\\
HR \hspace{-1.5mm} &
EDSR-B~\cite{lim2017edsr} \hspace{-1.5mm} &
CARN~\cite{ahn2018carn} \hspace{-1.5mm} &
IMDN~\cite{hui2019imdn} \hspace{-1.5mm} &
LatticeNet~\cite{luo2020latticenet} \hspace{-1.5mm} 
\\
\includegraphics[width=0.149\textwidth]{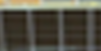} \hspace{-1.5mm} &
\includegraphics[width=0.149\textwidth]{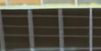} \hspace{-1.5mm} &
\includegraphics[width=0.149\textwidth]{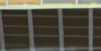} \hspace{-1.5mm} &
\includegraphics[width=0.149\textwidth]{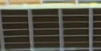} \hspace{-1.5mm} &
\includegraphics[width=0.149\textwidth]{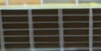} \hspace{-1.5mm}  
\\ 
Bicubic \hspace{-1.5mm} &
SwinIR-L~\cite{liang2021swinir} \hspace{-1.5mm} &
SRFormer-L~\cite{Zhou_2023srformer}  \hspace{-1.5mm} &
\textbf{HiT-SIR }(Ours)  \hspace{-1.5mm} &
\textcolor{black}{\textbf{HiT-SRF }}(Ours) \hspace{-1.5mm}
\\
\end{tabular}
\end{adjustbox}
\\
% % one row
\hspace{-0.2cm}
\begin{adjustbox}{valign=t}
\begin{tabular}{c}
\includegraphics[width=0.219\textwidth]{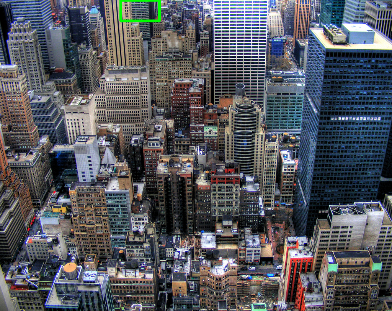}
\\
Urban100: img\_073 
\end{tabular}
\end{adjustbox}
\hspace{-0.23cm}
\begin{adjustbox}{valign=t}
\begin{tabular}{cccccc}
\includegraphics[width=0.149\textwidth]{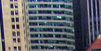} \hspace{-1.5mm} &
\includegraphics[width=0.149\textwidth]{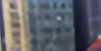} \hspace{-1.5mm} &
\includegraphics[width=0.149\textwidth]{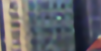} \hspace{-1.5mm} &
\includegraphics[width=0.149\textwidth]{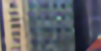} \hspace{-1.5mm} &
\includegraphics[width=0.149\textwidth]{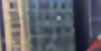} \hspace{-1.5mm} 
\\
HR \hspace{-1.5mm} &
EDSR-B~\cite{lim2017edsr} \hspace{-1.5mm} &
CARN~\cite{ahn2018carn} \hspace{-1.5mm} &
IMDN~\cite{hui2019imdn} \hspace{-1.5mm} &
LatticeNet~\cite{luo2020latticenet} \hspace{-1.5mm} 
\\
\includegraphics[width=0.149\textwidth]{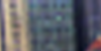} \hspace{-1.5mm} &
\includegraphics[width=0.149\textwidth]{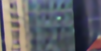} \hspace{-1.5mm} &
\includegraphics[width=0.149\textwidth]{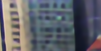} \hspace{-1.5mm} &
\includegraphics[width=0.149\textwidth]{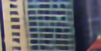} \hspace{-1.5mm} &
\includegraphics[width=0.149\textwidth]{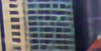} \hspace{-1.5mm}  
\\ 
Bicubic \hspace{-1.5mm} &
SwinIR-L~\cite{liang2021swinir} \hspace{-1.5mm} &
SRFormer-L~\cite{Zhou_2023srformer}  \hspace{-1.5mm} &
\textbf{HiT-SIR }(Ours)  \hspace{-1.5mm} &
\textcolor{black}{\textbf{HiT-SRF }}(Ours) \hspace{-1.5mm}
\\
\end{tabular}
\end{adjustbox}
\\
% % one row ----------------------------
\hspace{-0.2cm}
\begin{adjustbox}{valign=t}
\begin{tabular}{c}
\includegraphics[width=0.219\textwidth]{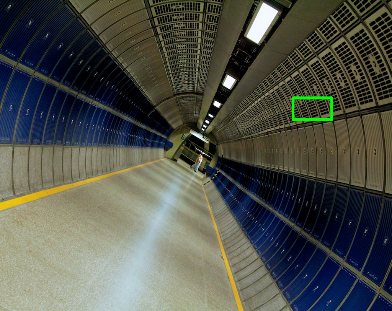}
\\
Urban100: img\_078 
\end{tabular}
\end{adjustbox}
\hspace{-0.23cm}
\begin{adjustbox}{valign=t}
\begin{tabular}{cccccc}
\includegraphics[width=0.149\textwidth]{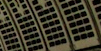} \hspace{-1.5mm} &
\includegraphics[width=0.149\textwidth]{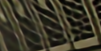} \hspace{-1.5mm} &
\includegraphics[width=0.149\textwidth]{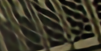} \hspace{-1.5mm} &
\includegraphics[width=0.149\textwidth]{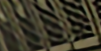} \hspace{-1.5mm} &
\includegraphics[width=0.149\textwidth]{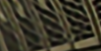} \hspace{-1.5mm} 
\\
HR \hspace{-1.5mm} &
EDSR-B~\cite{lim2017edsr} \hspace{-1.5mm} &
CARN~\cite{ahn2018carn} \hspace{-1.5mm} &
IMDN~\cite{hui2019imdn} \hspace{-1.5mm} &
LatticeNet~\cite{luo2020latticenet} \hspace{-1.5mm} 
\\
\includegraphics[width=0.149\textwidth]{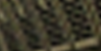} \hspace{-1.5mm} &
\includegraphics[width=0.149\textwidth]{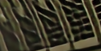} \hspace{-1.5mm} &
\includegraphics[width=0.149\textwidth]{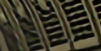} \hspace{-1.5mm} &
\includegraphics[width=0.149\textwidth]{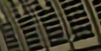} \hspace{-1.5mm} &
\includegraphics[width=0.149\textwidth]{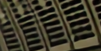} \hspace{-1.5mm}  
\\ 
Bicubic \hspace{-1.5mm} &
SwinIR-L~\cite{liang2021swinir} \hspace{-1.5mm} &
SRFormer-L~\cite{Zhou_2023srformer}  \hspace{-1.5mm} &
\textbf{HiT-SIR }(Ours)  \hspace{-1.5mm} &
\textcolor{black}{\textbf{HiT-SRF }}(Ours) \hspace{-1.5mm}
\\
\end{tabular}
\end{adjustbox}
\\
% % one row ----------------------------
\hspace{-0.2cm}
\begin{adjustbox}{valign=t}
\begin{tabular}{c}
\includegraphics[width=0.219\textwidth]{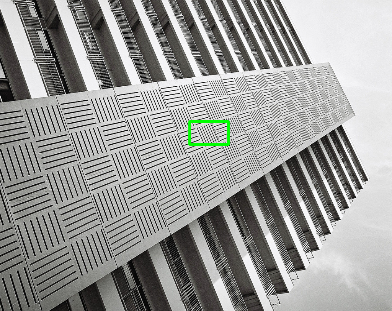}
\\
Urban100: img\_092 
\end{tabular}
\end{adjustbox}
\hspace{-0.23cm}
\begin{adjustbox}{valign=t}
\begin{tabular}{cccccc}
\includegraphics[width=0.149\textwidth]{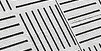} \hspace{-1.5mm} &
\includegraphics[width=0.149\textwidth]{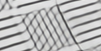} \hspace{-1.5mm} &
\includegraphics[width=0.149\textwidth]{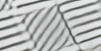} \hspace{-1.5mm} &
\includegraphics[width=0.149\textwidth]{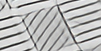} \hspace{-1.5mm} &
\includegraphics[width=0.149\textwidth]{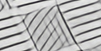} \hspace{-1.5mm} 
\\
HR \hspace{-1.5mm} &
EDSR-B~\cite{lim2017edsr} \hspace{-1.5mm} &
CARN~\cite{ahn2018carn} \hspace{-1.5mm} &
IMDN~\cite{hui2019imdn} \hspace{-1.5mm} &
LatticeNet~\cite{luo2020latticenet} \hspace{-1.5mm} 
\\
\includegraphics[width=0.149\textwidth]{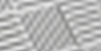} \hspace{-1.5mm} &
\includegraphics[width=0.149\textwidth]{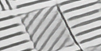} \hspace{-1.5mm} &
\includegraphics[width=0.149\textwidth]{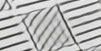} \hspace{-1.5mm} &
\includegraphics[width=0.149\textwidth]{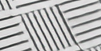} \hspace{-1.5mm} &
\includegraphics[width=0.149\textwidth]{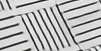} \hspace{-1.5mm}  
\\ 
Bicubic \hspace{-1.5mm} &
SwinIR-L~\cite{liang2021swinir} \hspace{-1.5mm} &
SRFormer-L~\cite{Zhou_2023srformer}  \hspace{-1.5mm} &
\textbf{HiT-SIR} (Ours)  \hspace{-1.5mm} &
\textcolor{black}{\textbf{HiT-SRF}} (Ours) \hspace{-1.5mm}
\\
\end{tabular}
\end{adjustbox}

\end{tabular} }
% \vspace{-1em}
\caption{{Qualitative comparisons for image SR ($\times$4) in challenging scenes. }}
\label{fig:qualires}
% \vspace{-3em}
\end{figure*}

\noindent \textbf{Quantitative Comparisons.}
% performance of CNN -> transformers -> our improvements (performance:tab1, efficiency:fig1, convergence: fig6), comparable S model with half param
% small model comparable with less param, compared with ori structure three aspect improvements
% overall performance, compared with original 
The results in Tab.~\ref{table:quantires} show the remarkable performance of our HiT-SR methods on all benchmark datasets. \textcolor{black}{Compared with the previous state-of-the-art approach SRFormer-Light, our HiT-SIR achieves comparable results with fewer parameters and FLOPs, and our HiT-SRF sets new state-of-the-art SR results across all upscaling factors.}
% Specifically, our small model HiT-SIR-S is able to achieve comparable results with the previous state-of-the-art approach SwinIR-NG, while reducing computational demands by nearly half.
Furthermore, our HiT-SR improved methods show advantages over their original models in three main aspects: performance, efficiency, and convergence. (i) As illustrated in Tab.~\ref{table:improvement}, our HiT-SR method contributes to significant improvements for SR performance, \textcolor{black}{\eg, 0.26/0.23/0.22 dB PSNR gains for HiT-SIR/HiT-SNG/HiT-SRF on Urban100 ($\times 2$) dataset.} (ii) By replacing the computationally inefficient shifted window self-attention with our SCC, all HiT-SR models require fewer computational resources and significantly boost inference speeds, achieving $\sim 7\times$ speed-up. Although the parameters and FLOPs improvements of HiT-SRF over SRFormer-Light are relatively smaller as SRFormer also improves the efficiency of transformer layers via permuted self-attention \cite{Zhou_2023srformer}, our HiT-SRF still shows better SR performance with $7.8\times$ faster inference speed, benefiting practical usage. (iii) We finally compare the convergence curves of SwinIR-Light and our improved HiT-SIR on Urban100 ($\times 2$) and Manga109 ($\times 2$) datasets in Fig.~\ref{fig:triancurve}. It is evident that our HiT-SR improved method converges faster than the original networks and achieves similar SR performance with only around half iterations.

\par 
\noindent \textbf{Qualitative Comparisons.}
% fig5, analyze recon details, failure: blurry details and artifacts 004 -> long range, distorted structure 076-078 -> hier, hierarchical features and long-range dependencies.
% We conduct qualitative comparisons in challenging scenes as depicted in Fig.~\ref{fig:qualires}.
Fig.~\ref{fig:qualires} shows qualitative comparisons in some challenging image SR scenes.
Previous SR approaches often suffer from blurry details and artifacts, \textcolor{black}{\eg, img\_078,} as they mainly utilize local features with limited receptive fields. In contrast, our HiT-SIR and HiT-SRF produce finer details and sharper textures by leveraging long-range dependencies with large windows. In addition, structure distortion is another common problem in SR results, especially in challenging scenarios like \textcolor{black}{img\_092.} We can observe that previous methods fail to recover the image structure and deteriorate SR performance with distorted lines. Instead, our HiT-SR methods utilize multi-scale information \eg, repetitive patterns, and better retain the overall structure. The effectiveness of our HiT-SR is further validated by comparing the visual results of \textcolor{black}{SwinIR-Light/SRFormer-Light and their improved versions HiT-SIR/HiT-SRF}, which restore better image structures with finer details.

\subsection{Method Analysis}
In this section, we further verify the advantages of our HiT-SR method by analyzing its information aggregation ability and window scalability.

\begin{wrapfigure}{r}{0.6\linewidth}
% \begin{figure*}[t]
% \vspace{-2em}
\centering
\begin{subfigure}{0.18\linewidth}
\captionsetup{justification=centering}
%     \includegraphics[width=\linewidth]{imgs/supp/LAM/1.png}
% \\
% \includegraphics[width=\linewidth]{imgs/supp/LAM/4.png}
% \\
\includegraphics[width=\linewidth]{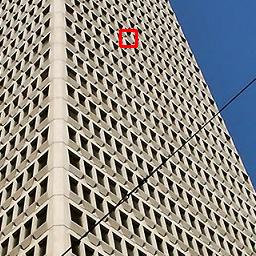}
\\
% \includegraphics[width=\linewidth]{imgs/supp/LAM/6.png}
% \\
\includegraphics[width=\linewidth]{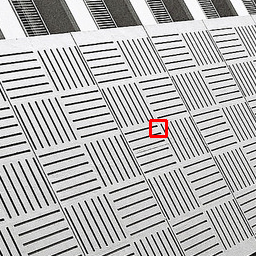}
    \caption*{\tiny Target \\ Region}
\end{subfigure}
%% new coloumn
\begin{subfigure}{0.18\linewidth}
\captionsetup{justification=centering}
% \includegraphics[width=\linewidth]{imgs/supp/LAM/SwinIR-white1.png}
% \\
% \includegraphics[width=\linewidth]{imgs/supp/LAM/SwinIR-white4.png}
% \\
\includegraphics[width=\linewidth]{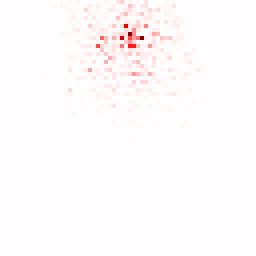}
\\
% \includegraphics[width=\linewidth]{imgs/supp/LAM/SwinIR-white6.png}
% \\
\includegraphics[width=\linewidth]{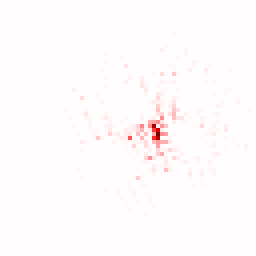}
    \caption*{\tiny LAM \\ (SIR)}
\end{subfigure}
%% new coloumn
\begin{subfigure}{0.18\linewidth}
\captionsetup{justification=centering}
%     \includegraphics[width=\linewidth]{imgs/supp/LAM/HITSIR-white1.png}
% \\
% \includegraphics[width=\linewidth]{imgs/supp/LAM/HITSIR-white4.png}
% \\
\includegraphics[width=\linewidth]{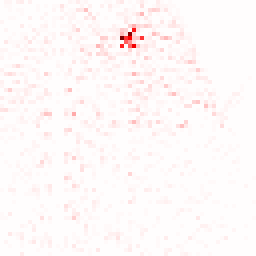}
\\
% \includegraphics[width=\linewidth]{imgs/supp/LAM/HITSIR-white6.png}
% \\
\includegraphics[width=\linewidth]{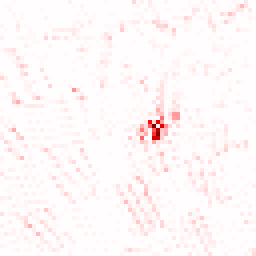}
    \caption*{\tiny LAM \\ (HiT-SIR)}
\end{subfigure}
%% new coloumn
\begin{subfigure}{0.18\linewidth}
\captionsetup{justification=centering}
%     \includegraphics[width=\linewidth]{imgs/supp/LAM/SwinIR-layer1.png}
% \\
% \includegraphics[width=\linewidth]{imgs/supp/LAM/SwinIR-layer4.png}
% \\
\includegraphics[width=\linewidth]{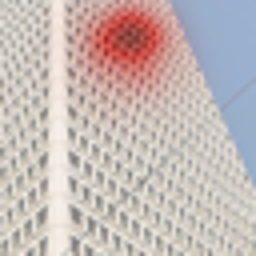}
\\
% \includegraphics[width=\linewidth]{imgs/supp/LAM/SwinIR-layer6.png}
% \\
\includegraphics[width=\linewidth]{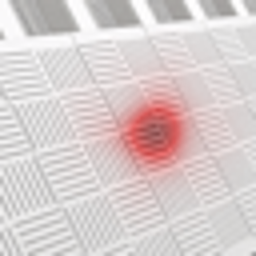}
    \caption*{\tiny Info. Area \\ (SIR)}
\end{subfigure}
%% new coloumn
\begin{subfigure}{0.18\linewidth}
\captionsetup{justification=centering}
%     \includegraphics[width=\linewidth]{imgs/supp/LAM/HITSIR-layer1.png}
% \\
% \includegraphics[width=\linewidth]{imgs/supp/LAM/HITSIR-layer4.png}
% \\
\includegraphics[width=\linewidth]{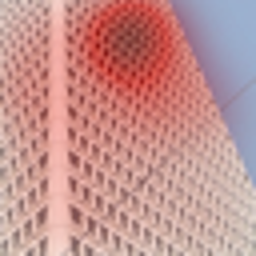}
\\
% \includegraphics[width=\linewidth]{imgs/supp/LAM/HITSIR-layer6.png}
% \\
\includegraphics[width=\linewidth]{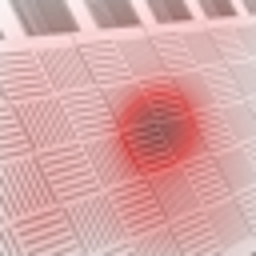}
    \caption*{\tiny Info. Area \\ (HiT-SIR)}
\end{subfigure}
% \vspace{-1em}
        % \caption{Local attribution maps (LAM) and informative areas (Info. Area) of SwinIR-Light (SIR) and HiT-SIR. }
	\caption{Comparisons between SwinIR-Light (SIR) and HiT-SIR, where local attribution maps (LAM) and informative areas (Info. Area) are displayed. }
	\label{fig:lam}
 % \vspace{-1.5em}
\end{wrapfigure}

\par 
\noindent \textbf{Information Aggregation.} We employ the local attribution maps (LAM) \cite{gu2021interpreting} to analyze the information aggregation performance of HiT-SR. LAM is a visual tool aiming at finding input pixels that strongly influence the SR outputs, \ie, informative areas, and larger informative areas mean better aggregation ability. 
% Thus, by analyzing the LAM results for a target region, we can compare the information aggregation performance of SR methods, and larger informative areas mean better aggregation ability. 
As shown in Fig.~\ref{fig:lam}, we apply the LAM approach to SwinIR-Light \cite{liang2021swinir} and our improved version HiT-SIR, and compare the informative areas for the same target regions. Due to the fixed small windows used in SwinIR-Light, the informative areas are limited to a local region, hindering the utilization of long-range dependencies. By contrast, our HiT-SIR significantly expands the informative areas by leveraging hierarchical features. Therefore, our proposed method is able to gather a wider range of information, \eg, similar structures and repetitive patterns, to boost SR performance.

\begin{wrapfigure}{r}{0.45\linewidth}
\centering
% \vspace{-2em}
       \begin{tikzpicture}[scale=1]
		\begin{axis}
			[
             height=0.75\linewidth,
             width=\linewidth,
             ymin=0, ymax=100,
             % xmode=log,
             ymode=log,
              ylabel={\textbf{Log-Scale FLOPs (G)}},
              xlabel={Window size (pixel)},
              xtick={16,32,64,96,128},
              ytick={0.01,0.1,1,10,100},
              line width=0.8pt,
			 axis y line=left,
			 axis x line=bottom,
              every axis/.append style={font=\tiny},
              grid,
              legend style={fill=none,draw=none,font=\tiny, at={(0.02,0.98)},anchor=north west},
    grid style={color=white, line width=0.5pt},
              axis background/.style={fill=gray!10}, % 设置背景颜色
			 enlarge x limits=0.1, ] 
			\addplot+ [mark=o] table [col sep=comma, x=Step, y=Value] {data/main/WSA-flops.csv} 
   node[pos=0.428, anchor=south] {2.1}
   node[pos=0.714, anchor=south] {10.3}
   node[pos=1, anchor=south] {32.4};
                \addlegendentry{W-SA}
                \addplot+ [mark=square] table [col sep=comma, x=Step, y=Value] {data/main/PSA-flops.csv} 
                node[pos=0.428, anchor=south] {0.5}
                node[pos=0.714, anchor=south] {2.6}
                node[pos=1, anchor=south] {8.2};
                \addlegendentry{PSA}
                \addplot+ [black,mark=triangle] table [col sep=comma, x=Step, y=Value] {data/main/SCC-flops.csv} 
                node[pos=0.428, anchor=south] {0.1}
                node[pos=0.714, anchor=south] {0.2}
                node[pos=1, anchor=south] {0.3};
                \addlegendentry{\textbf{SCC}}
		\end{axis} 
	\end{tikzpicture} 
    % \vspace{-1em}
    \caption{FLOPs of W-SA in SwinIR~\cite{liang2021swinir}, PSA in SRFormer~\cite{Zhou_2023srformer}, and our SCC, with respect to square window sizes.}
    % \caption{FLOPs of W-SA \cite{liang2021swinir}, PSA  \cite{Zhou_2023srformer}, and SCC, \wrt square window sizes.}
    \label{fig:flops}
    % \vspace{-1.5em}
\end{wrapfigure}
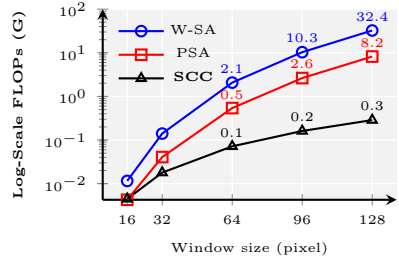
\par 
\noindent \textbf{Window Scalability.} We measure the required FLOPs for a single W-SA \cite{liang2021swinir}, PSA \cite{Zhou_2023srformer}, and our SCC layer, under the same hyperparameter settings (channel and head numbers are set to 60 and 6, respectively). As depicted in Fig.~\ref{fig:flops}, the FLOPs of W-SA increase drastically with the expansion of window size, making it unaffordable to utilize large windows. Although PSA alleviates the computational burden with permutation techniques, its complexity still remains quadratic to window sizes, limiting window scaling. Benefiting from the linear complexity of SCC, our method equips SR transformers with scalable large windows to efficiently gather long-range information. 

% can easily utilize large windows
% By contrast, our SCC easily supports large windows
% To test the sca

\subsection{Ablation Study}
In Tab.~\ref{table:ablation}, we study the effectiveness of each component in our design based on HiT-SIR network. All models are trained on the DIV2K dataset \cite{agustsson2017ntire_div2k} for 300K iterations and tested on the Manga109 ($\times 2$) dataset \cite{matsui2017sketch_manga109}. We set the output size to $3\times 720\times 1280$ to compute FLOPs. Note that we maintain similar complexity across all variants by adjusting channel numbers for fair comparisons.

\noindent \textbf{Hierarchical Windows.}
% fixed small and large not good, all large computational expensive, hierarchical good, expanding good 
To verify the effect of our window strategy, we conduct experiments with different window arrangement methods, including fixed small windows (\#2), fixed large windows (\#3), and shrinking hierarchical windows (\#4). The model with small windows can only utilize local features for SR, resulting in sub-optimal performance. By establishing long-range dependencies, \#3 boosts the PSNR results but simultaneously brings more computational burdens due to the fixed large window sizes. Compared with them, the models with hierarchical windows (\#1 and \#4) make full use of multi-scale features for SR, showing promising improvements in both performance and efficiency. Moreover, the proposed expanding hierarchical windows gain the best performance. This is because our expanding window strategy allows networks to first utilize the most relevant information and improve feature representation with small windows, benefiting the establishment of reliable long-range dependencies with the subsequent large windows.

\begin{table}[t]
\scriptsize
\caption{Ablation study. We train all models on DIV2K for 300K iterations, and test on Manga109 ($\times 2$). The ablations about hierarchical windows, DFE, SCC, and different head strategies correspond to \#1-4, \#5-7, \#8-12, and \#13-15, where our strategies are noted in \textbf{bold}. Win. indicates the assigned window sizes for a transformer block, and \#head means the number of correlation heads. Best results are colored \textcolor{red}{red}.}
\label{table:ablation}
% \vspace{-2.5em}
\begin{center}
\setlength\tabcolsep{5pt}
\begin{tabular}{c|c|cc|cc}
\toprule[0.15em]
\rowcolor{color3}
ID   & Strategies                       & \#Para. & FLOPs & PSNR & SSIM \\ 
\midrule[0.15em]
\#1  & \textbf{Win.={[}4,8,16,32,48,64{]}}  & 772K         & 209.9G      &  \textcolor{red}{39.21}    &  \textcolor{red}{0.9780}    \\
\#2  & Win.={[}8,8,8,8,8,8{]}       & 772K         & 208.6G      & 38.88     &  0.9775    \\
\#3  & Win.={[}64,64,64,64,64,64{]} & 773K         & 241.4G      &  39.09    & 0.9775     \\
\#4  & Win.={[}64,48,32,16,8,4{]}   & 772K         & 209.9G      & \textcolor{black}{39.12}     &  \textcolor{black}{0.9779}    \\
\midrule
\#5  & \textbf{DFE}                         &   772K       & 209.9G      &  \textcolor{red}{39.21}    &  \textcolor{red}{0.9780}   \\
\#6  & DFE $\rightarrow$ Linear     &  792K       & 214.9G      &  \textcolor{black}{38.99}    &  \textcolor{black}{0.9777}    \\
\#7  & DFE $\rightarrow$ Conv.      &  781K        &    212.3G   &   38.95   & 0.9775     \\ 
\midrule
\#8 & \textbf{SCC}                         &  772K        & 209.9G      &  \textcolor{red}{39.21}    &  \textcolor{red}{0.9780}  \\
\#9 & QV $\rightarrow$ QKV         &  826K        & 222.6G      &  \textcolor{black}{39.19}    &  \textcolor{red}{0.9780}    \\
\#10 & Correlation $\rightarrow$ Attention    &  772K        & 209.9G     & 39.08     &  \textcolor{black}{0.9779}    \\
\#11 & S-SC only       &  819K        & 222.1G      &   39.04   &  0.9775    \\
\#12 & C-SC only        &  816K        & 217.2G      &  39.18    &  \textcolor{black}{0.9778}    \\ 
\midrule
\#13 & \textbf{\#head of S-SC, C-SC=[6,1]}  &  772K        &  209.9G     &  \textcolor{red}{39.21}    &  \textcolor{red}{0.9780}  \\
\#14 & \#head of S-SC, C-SC=[1,1]   &  772K        & 209.9G      &  \textcolor{black}{39.14}    &  \textcolor{black}{0.9778}    \\
\#15 & \#head of S-SC, C-SC=[6,6]   &  772K        &  201.2G     &   39.08   & \textcolor{black}{0.9778}    \\
\bottomrule[0.15em]
\end{tabular}

\end{center}
% \vspace{-3em}
\end{table}

\par 
\noindent \textbf{Dual Feature Extraction.}
In the layer-level design, we first investigate the impact of DFE by comparing it with linear projection (\#6) and convolutional projection (\#7). As shown in Tab.~\ref{table:ablation}, the performance of linear projection is limited since only channel information is leveraged. Although spatial relation is exploited in convolution, efficiency-related techniques like dimension reduction or depth-wise operation are often employed to alleviate the computational burdens of convolutional layers, leading to insufficient utilization of channel information. By contrast, our DFE extracts spatial and channel information in a two-branch structure and fuses features from different domains, gaining significant performance improvements as shown in Tab.~\ref{table:ablation}.

\par 
\noindent \textbf{Spatial-Channel Correlation.}
% QV-> QKV information redundancy; corr2attn better QV with correlation transforming matrix directly functions on values to build more accurate XXX; information aggregation, 
Our SCC efficiently aggregates spatial and channel information via self-correlation with queries and values. We first design a variant using commonly adopted queries, keys, and values for computation (QKV in \#9). As illustrated in Tab.~\ref{table:ablation}, the QKV setting performs similarly to our QV setting but consumes more computational resources. This is because keys and values both reflect the intrinsic properties of input features, and thus using only queries and values mitigates the information redundancy during computation. In addition, involving values in computing the correlation matrix, \ie, Eq.~\eqref{eq:ssc} and \eqref{eq:csc}, also contributes to SR performance with more precise transforming relations. Next, we add softmax layers to Eq.~\eqref{eq:ssc} and \eqref{eq:csc}, converting our correlation methods to attention approaches (\#10) for comparison. The results in Tab.~\ref{table:ablation} validate the better performance of our proposed methods without limiting the transformation scales by softmax operations.
Finally, we conduct experiments by replacing SCC with only S-SC (\#11) or C-SC (\#12). From the metrics in Tab.~\ref{table:ablation}, we can observe that C-SC performs better than S-SC, and utilizing both spatial and channel information leads to the overall best performance.

\par 
\noindent \textbf{Different Correlation Head.}
% single head for spatial cause drop, multi-head for channel cause drop
To study our different correlation head strategy, we train two models with 
the same single-head and multi-head strategies for both S-SC and C-SC, corresponding to \#14 and \#15 in Tab.~\ref{table:ablation}. By comparing \#13 and \#14, one can see that adopting the single-head method for S-SC is inappropriate as single-head prevents S-SC from aggregating features from different representation spaces. Meanwhile, applying the multi-head approach to C-SC, \ie, \#15, also impairs SR performance since splitting the channel into different heads prohibits each channel from gathering information from all the other channels. Compared with these variants, our different correlation head strategy achieves the best results by enabling S-SC to aggregate features from different channel subspaces while allowing full channel interaction in C-SC.
% the full potential of channel correlation for C-SC.

\section{Conclusion}
In this paper, we propose a general strategy to convert popular transformer-based SR methods to hierarchical transformers for efficient image SR (HiT-SR). Our approach consists of block-level and layer-level designs. In each transformer block, we apply expanding hierarchical windows to establish long-range dependencies and leverage multi-scale features, boosting SR performance. Considering the quadratic complexity of self-attention methods, we devise a spatial-channel correlation (SCC) method with linear complexity to window sizes, benefiting the efficient aggregation of hierarchical features. 
Extensive evaluations are made to verify the effectiveness of the proposed HiT-SR, and our HiT-SIR, HiT-SNG, and HiT-SRF set new state-of-the-art results for efficient image SR.

% In this paper, we propose a general strategy to convert popular transformer-based SR methods to hierarchical transformers for efficient image SR (HiT-SR). Our approach consists of block-level and layer-level designs. In each transformer block, we apply expanding hierarchical windows to establish long-range dependencies and leverage multi-scale features, boosting SR performance. Considering the computational burden of self-attention methods in processing large windows, we devise a spatial-channel correlation (SCC) method with linear complexity to window sizes, benefiting the efficient aggregation of hierarchical features. In addition, our SCC takes into account both spatial and channel information, achieving performance improvements while maintaining efficiency. 
% Extensive evaluations are made to verify the effectiveness of the proposed HiT-SR, and our HiT-SIR, HiT-SNG, and HiT-SRF set new state-of-the-art results for efficient image SR.

\section*{Acknowledgements}
This work was supported by Shanghai Municipal Science and Technology Major Project (2021SHZDZX0102), the Fundamental Research Funds for the Central Universities, and Huawei Technologies Oy (Finland) Project.

\bibliographystyle{splncs04}
\bibliography{main}
\end{document}